\documentclass{elsarticle}
\usepackage[centertags]{amsmath}
\usepackage{amsthm}
\usepackage{bm}
\usepackage{units}
\usepackage{amssymb}
\usepackage{caption}
\usepackage[caption=false]{subfig}

\DeclareMathAlphabet{\mbf}{OT1}{ptm}{b}{n}

\usepackage{amssymb}
\usepackage{graphicx}
\usepackage{varioref}
\labelformat{equation}{(#1)}

\usepackage[margin=2.5cm]{geometry}

\journal{Neural Networks}

\usepackage{natbib}
\biboptions{authoryear}

\def\bfE{{\mbf E}}
\def\f{f}
\newcommand{\ff}{{f\!f}}
\def\bfH{{\mbf H}}
\def\bfK{{\mbf K}}
\def\bfW{{\mbf W}}
\def\bfq{{\mbf q}}
\def\qdot{\dot{\mbf q}}
\def\bfom{\bm{\omega}}
\def\bfet{\bm{\eta}}
\def\Rtilde{\tilde{R}}
\def\Re{\mathbb{R}}
\def\Stilde{{\tilde{S}}}
\def\bfx{{\mbf x}}
\def\xdot{\dot{\bfx}}
\def\xtilde{\tilde{\mbf x}}

\def\bfz{{\mbf z}}
\def\ztilde{\tilde{\mbf z}}
\def\Zero{{\mbf O}}
\def\x{\times}
\def\bfA{{\mbf A}}
\def\bfB{{\mbf B}}
\def\bfC{{\mbf C}}

\newtheorem{thm}{Theorem}
\newtheorem{pf}{Proof}

\begin{document}

\begin{frontmatter}

\title{Synthesis of recurrent neural networks for dynamical system simulation}

\author[maluubaaddress]{Adam P Trischler\corref{corauth}}
\cortext[corauth]{Corresponding author.}
\ead{adam.trischler@maluuba.com}

\author[utiasaddress]{Gabriele MT D'Eleuterio}
\ead{gabriele.deleuterio@utoronto.ca}

\address[maluubaaddress]{Maluuba Research, 2000 Peel Street, Montreal}
\address[utiasaddress]{University of Toronto Institute for Aerospace Studies, 4925 Dufferin Street, Toronto}

\begin{abstract}
We review several of the most widely used techniques for training recurrent neural networks to approximate dynamical systems, then describe a novel algorithm for this task. The algorithm is based on an earlier theoretical result that guarantees the quality of the network approximation. We show that a feedforward neural network can be trained on the vector-field representation of a given dynamical system using backpropagation, then recast it as a recurrent network that replicates the original system's dynamics. After detailing this algorithm and its relation to earlier approaches, we present numerical examples that demonstrate its capabilities. One of the distinguishing features of our approach is that both the original dynamical systems and the recurrent networks that simulate them operate in continuous time.
\end{abstract}

\begin{keyword}
recurrent neural network\sep dynamical system\sep approximation\sep attractor \sep chaos \sep nonautonomous system
\MSC[2010] 00-01\sep  99-00
\end{keyword}

\end{frontmatter}

\section{Introduction and survey}
Recurrent neural networks (RNNs) represent a large class of computational models designed in analogy to the brain. What distinguishes them from the better known feedforward neural networks is the existence of closed cycles in the connection topology; as a consequence of these cycles, RNNs may exhibit self-sustained dynamics in the absence of any input \citep{jaeger2009}. Mathematically, RNNs are dynamical systems.

Not only is the brain characterized by massively recurrent connectivity, but dynamical systems themselves are now a mainstay of computational neuroscience. Persistent activity in biological neural networks is posited to result from dynamical attractors in neural state space \citep{amit1992}, and dynamical computation underlies a variety of models for information processing and memory function in the brain \citep{eliasmith2004,kaneko2003,afraimovich2004}.

\subsection{RNN training techniques}
Motivated by these facts and other interests, various researchers have developed techniques by which recurrent networks can be trained to approximate dynamical systems.

One set of such techniques takes a discrete-time approach: dynamical systems, whether continuous or discrete in time, are approximated with discrete-time RNNs. This set includes backpropagation through time \citep{werbos1990}, real-time recurrent learning \citep{williams1989}, the extended Kalman filter \citep{feldkamp1998}, reservoir computing \citep{jaeger2004}, and phase-space learning~\citep{tsung1995}.

Backpropagation through time (BPTT) adapts the standard backpropagation algorithm~\citep{rumelhart1988} for feedforward networks to recurrent networks. BPTT works by ``unfolding'' a network in time: identical copies of the RNN are stacked in layers, and connections within the network are redirected to obtain connections between subsequent copies. Each layer represents the same network at a different step in time. The result of this unfolding is a feedforward network amenable to standard backpropagation. BPTT is probably the most widely used method for RNN training and can be made to perform very well with various modifications ({\it e.g.}, stochastic sample selection~\citep{bottou2010} and the addition of measures for computational efficiency and stability like momentum~\citep{sutskever2013}). One common issue is that error gradients shrink or expand exponentially over time because they are multiplied repeatedly by copies of the same weight matrix. This is referred to as the problem of vanishing and exploding gradients. As a consequence, long-term memory effects are quite difficult to train. This issue can be overcome with second-order optimization techniques that use curvature information, such as the popular Hessian-free~\citep{martens2011}, or by modifiying the network architecture. The Long Short-Term Memory (LSTM) network, for example, features multiplicative gates and linear units whose gradients are unity so that their values can be remembered over many timesteps~\citep{hochreiter1997}.

Real-time recurrent learning (RTRL) computes a RNN's exact error gradient at every discrete timestep~\citep{williams1989}. Differentiating the network equations by the weights yields a discrete-time linear dynamical system with time-varying coefficients, where the resulting partial derivatives of the network state are the dynamical variables. Iterating through time for the error gradients simultaneously with the network dynamics provides the required weight adjustment. Although RTRL is mathematically transparent, the computational cost for each update step is $O(n^4)$. This renders the scheme practicable only for very small networks on the order of about ten neurons~\citep{jaeger2009}.

The extended Kalman filter (EKF) is a state estimation technique for nonlinear systems derived by linearizing the original Kalman filter about the current state estimate. It is a second-order gradient-descent algorithm that uses curvature information from the squared error surface. The pioneering work in adapting the EKF to RNN training was done by~\cite{feldkamp1998} within the domain of system identification. Here, the unknown network weights are interpreted as the state of a dynamical system, and the desired dynamical trajectory as a measurement of that state.

In reservoir computing (RC), a randomly recurrently connected sea of neuron units, referred to as the {\it reservoir}, feeds forward to a set of output units and may be driven by an input signal. The reservoir functions as a dynamical system and must exhibit what is called the {\it echo state property}. This relates asymptotic properties of the reservoir dynamics to the driving signal. Intuitively speaking, the echo state property means that the reservoir asymptotically eliminates any information from initial conditions~\citep{jaeger2009}. The output units in RC may be made to approximate prescribed dynamical trajectories by training only the output weights that feed them (typically by a linear regression process); the random recurrent connections within the reservoir remain fixed. An issue with RC is that the random reservoir is, to date, poorly understood from a theoretical standpoint \citep{jaeger2009}. Networks generated by our procedure in fact share a structural relation to RC systems, which we shall discuss. 

The phase-space learning method (PSL) of \cite{tsung1995} takes a vector-field approach to RNN training that is similar to our algorithm from a high-level perspective. PSL consists of (1) embedding a dynamical trajectory to recover its phase-space structure (this is an application of~\cite{takens1981} theorem); (2) generating local approximations of the underlying vector field about the given trajectory; and (3) approximating the vector field with a feedforward network. This method transforms the recurrent network problem into a feedforward one, as does ours. However, the networks generated by our approach and that of \cite{tsung1995} differ significantly. PSL networks are discrete in time, and they also remain essentially feedforward even after training. They do not contain a recurrently connected reservoir of hidden neurons; recurrence only arises in piping network outputs back to the input neurons.

As noted, these techniques train discrete-time RNNs.\footnote{A continuous-time EKF exists, but its application to the RNN training problem requires continuous-time derivatives for the Jacobian.}\footnote{\cite{hermans2010} adapted reservoir computing to continuous time.} Discrete-time networks are simpler conceptually and easier to train, and of course numerical simulations are carried out on discrete-time digital computers. However, discrete-time systems are of less interest to neuroscience than their continuous-time counterparts since the brain is inherently continuous.\footnote{Although the brain operates on spike trains, the spikes themselves are best modelled by ODEs, such as the Hodgkin-Huxley model.} It is also well known that finite-difference equations can behave very distinctly from ODEs in some respects: {\it e.g.}, chaos can occur in one-dimensional finite-difference systems, while three dimensions are required for chaos in continuous systems. This, along with the desire to apply tools of functional analysis, motivated us to set our RNNs within a continuous-time formalism. In our work, both the original dynamical system and its recurrent-network approximation are modeled with ordinary differential equations (ODEs).

To our knowledge, the most widely used technique for training continuous-time recurrent networks is the Neural Engineering Framework (NEF) of \cite{eliasmith2004, eliasmith2005}. In this framework, the activities of a population of neurons encode some input vector. Given this encoding, the original stimulus vector can be approximately recovered by decoding the population activities; this is accomplished by taking the inner product of the activities with a set of decoding vectors. Decoding vectors are determined by a least-squares method. Similarly, approximate transformations of the original stimulus vectors can be defined using a modified set of decoding vectors, referred to as transformational decoders.

An interesting aspect of the NEF that bears some relation to the RC approach is that the linear encoder weights are set randomly; they are not trained. Contrary to typical backpropagation, error is evaluated only at the output, based on the linear decoders; it is not propagated back to assign credit or blame to the linear encoders. In some ways this is advantageous: because errors need not propagate through the neuron activation function, this function need not be differentiable. In the NEF, both the activation function and the transfer functions of its neurons are generic. For example, the activation function can be continuous (sigmoidal) or spiking (leaky integrate-and-fire).

There exist strong analogies between the NEF and our RNN training technique, although the two methods spring from different formalisms. We take as our starting point a theorem from the continuous-time RNN literature that is not employed by the NEF authors. Through it, we arrive at a procedure and a class of networks which can be viewed as a special case of the NEF. However, by starting from this theorem and tracing a different route to the end procedure, we show that the special case of networks we utilize is governed by theoretical bounds on its performance. To our knowledge, this theoretical support for (a special case of) the NEF was not known previously. We will characterize the relation to the NEF mathematically in \S\ref{sec:disc}, after detailing our procedure in \S\ref{sec:rnn}.

\subsection{The proposed RNN training procedure}
Our algorithm for training RNNs to approximate prescribed dynamical systems is based on a theoretical result of \cite{funahashi1993}. Theorem 1 therein states that any dynamical system can be ``approximated to arbitrary accuracy'' by a recurrent neural network. This theorem is an existence result; here, we will present a constructive algorithm for obtaining the approximating networks that \cite{funahashi1993} theorize.

In our approach, a feedforward neural network is first trained on the vector-field representation of a given dynamical system using standard backpropagation techniques. Then the trained feedforward network is recast, using matrix manipulations, as a continuous-time recurrent network that replicates the original system's dynamics. Using the result of \cite{funahashi1993}, we provide an upper bound on the constructed RNN's trajectory error as a function of the error in feedforward training. Because we use shallow feedforward networks, the vanishing/exploding gradients problem is circumvented.

We also show how recurrent networks trained in this manner can be coupled additively, and prove that coupled systems achieve arbitrary accuracy as a function of the training error. This enables us to decompose certain dynamical systems, train on their simpler subcomponents, and then combine the constructed RNNs to simulate the full system. We will demonstrate this method specifically by the simulation of driven dynamical systems ({\it i.e.,} systems with external inputs). Driven systems were not handled by the theory of \cite{funahashi1993}, although \cite{chow2000} later made this extension. Our approach for simulating driven systems is different from that of \cite{chow2000} in that we simulate the driving signal, as well as the system upon which the driving acts, with recurrent neural networks. On the other hand, \cite{chow2000} take the driving signal as given and show how this signal can be connected appropriately to the driven system (we present the details in Section~\ref{sec:rnn}).

The theory of \cite{chow2000} admits arbitrary bounded input signals, and is therefore stronger than our result, which obtains only in the special case that the driving signal is itself a variable of an autonomous dynamical system and where the driving input is additive. But given our motivation (decomposing dynamical systems and coupling RNNs) we seek to capture everything, including the driving signal, within a recurrent network. Furthermore, \cite{chow2000} never demonstrate that their theorem can be harnessed constructively to learn forced recurrent networks from data.

The remainder of this paper is organized as follows. In Section~\ref{sec:rnn}, we present the mathematics for synthesis of recurrent neural networks from feedforward networks, along with the training methodology and the means for simulating driven dynamical systems. Section~\ref{sec:res} details several examples of dynamical systems realized as RNNs while highlighting potential applications. In Section~\ref{sec:disc}, we present an analysis of certain network parameters and show how potential functions for our networks can be determined; in that section the correspondence between reservoir computing and RNNs generated by our algorithm is related, as is the relation to the NEF. In Section~\ref{sec:conc}, we offer a few concluding remarks.

\section{Synthesis of recurrent neural networks}
\label{sec:rnn}

\subsection{Theoretical background}
\label{sub:ff-rnn}

Our method for training a recurrent neural network to replicate a dynamical system is based on the following theorem.
\begin{thm}[\cite{funahashi1993}]
Let $D$ be an open subset of ${}^n\mathbb{R}$,\footnote{We indicate the space of $p\x q$ real matrices as ${}^p\mathbb{R}^q$ and accordingly ${}^p\mathbb{R}$ is the space of $p\x 1$ real column matrices.}
$F: D \rightarrow {}^n\mathbb{R}$ be a $C^1$-mapping, and $\tilde{K}$ be a compact subset of $D$. Suppose that there is a subset $K \subset \tilde{K}$ such that any solution $\bfq(t)$ with initial value $\bfq(0)$ in $K$ of an ordinary differential equation
\begin{equation}
\dot{\bfq} = F(\bfq), \qquad \bfq(0) \in K
\label{eq:fn1}
\end{equation}
is defined on $I = [0, T]$, $0 < T < \infty$ and $\bfq(t)$ is included in $\tilde{K}$ for any $t \in I$. Then, for an arbitrary $\epsilon > 0$, there exist an integer $m$ and a recurrent neural network with $n$ output units and $m$ hidden units such that for a solution $\bfq(t)$ satisfying~\ref{eq:fn1} and an appropriate initial state of the network,
\begin{equation}
\max_{t \in I} \| \bfq(t) - \bfom(t) \| < \epsilon
\label{eq:fn2}
\end{equation}
holds, where $\bfom(t) = (\omega_1(t), \ldots, \omega_n(t))^T$ is the internal state of the output units of the network.
\label{thm1}
\end{thm}

The required recurrent network has $n$ output units, whose states replicate the $n$-dimensional orbits of the prescribed dynamical system, and $m$ hidden units. Its particular form, plus initial condition, is
\begin{equation}
    \dot{\mbf s} = G({\mbf s}) \triangleq -\frac{1}{\tau}{\mbf s} + {\mbf W}\bm{\sigma}({\mbf s}),\qquad 
    {\mbf s}(0) = \left(\begin{array}{c} 
        {\mbf q}(0) \\ {\mbf B}{\mbf q}(0)+\bm{\theta}
        \end{array}\right),
\label{eq:G}
\end{equation}
where $G:{}^{n+m}\mathbb{R}\rightarrow{}^{n+m}\mathbb{R}$, ${\mbf s}(t) \in {}^{n+m}\mathbb{R}$ is the vector of internal neuron states, $\tau$ is the neuron time constant, ${\mbf W} \in {}^{n+m}\mathbb{R}^{n+m}$ is the matrix of connection weights, and $\bm{\sigma}({\mbf x}) = \text{col}\,[\sigma(x_i)]$ is the sigmoidal activation function. We will describe the matrices ${\mbf W}$ and $\bfB$ and the vector $\bm{\theta}$ shortly.

The neurons consist of two sets: the output units, whose activations are given by $\bm{\omega}(t) \in {}^n\mathbb{R}$, and hidden units, whose activations are $\bm{\eta}(t) \in {}^m\mathbb{R}$, so that
\begin{equation}
    {\mbf s}
	   = \left( \begin{array}{c}
		  \bm{\omega} \\
		  \bm{\eta} \end{array} \right).
\label{eq:scat}
\end{equation}
For the activation function, the hyperbolic tangent or the logistic sigmoid, $\sigma(x) = 1/(1+\exp(-x))$, is typically used.

The weight matrix of the approximating RNN is given by
\begin{equation}
	{\mbf W}
	   = \left( \begin{array}{cc}
		  \Zero & {\mbf A} \\
		  \Zero & {\mbf B}{\mbf A} \end{array} \right)
            \in {}^{n+m}\mathbb{R}^{n+m}.
\label{eq:W}
\end{equation}
Here, the block ${\mbf A} \in {}^n\mathbb{R}^m$ represents connections from the hidden units to the output units. The block ${\mbf B}{\mbf A}$, with ${\mbf B} \in {}^m\mathbb{R}^n$, represents connections among hidden units---it is here that recurrent loops lie.

The key to Theorem~\ref{thm1}'s proof, and our training methodology, is determining the appropriate matrices ${\mbf A}$ and ${\mbf B}$. \cite{funahashi1993} show that their result holds if
\begin{equation}
    E_{\max} \triangleq \max_{\mbf q} \| F({\mbf q}) - {\mbf A}\bm{\sigma}({\mbf B}{\mbf q}+\bm{\theta}) \| < \frac{\epsilon L_F}{4(e^{L_F T}-1)},
\label{eq:vferror}
\end{equation}
where $L_F$ is the Lipschitz constant of $F$. The left-hand side of the inequality in~\ref{eq:vferror} can be interpreted as the maximum error over the domain of two vector fields: field $F$ describes the dynamical system we wish to approximate; the field
\begin{equation}
{\mbf A}\bm{\sigma}({\mbf B}{\mbf q}+\bm{\theta}) \triangleq F_\text{FF}(\bfq)
\label{eq:ffn}
\end{equation}
takes the form of a three-layer feedforward neural network. In particular, matrix ${\mbf B}$ represents the connections from the input units to the hidden units, vector $\bm{\theta} \in {}^m\mathbb{R}$ is the bias on the hidden units, and matrix ${\mbf A}$ represents the connections from the hidden units to the output units.

As a consequence of this special form, the earlier {\it fundamental approximation theorem} for neural networks~\citep{funahashi1989} can be invoked to show that the required ${\mbf A}$, ${\mbf B}$, and $\bm{\theta}$ exist.
\begin{thm}[\cite{funahashi1989}]
Let $K$ be a compact subset of ${}^n\mathbb{R}$ and $F: K \rightarrow {}^n\mathbb{R}$ be a continuous mapping. Then, for any arbitrary $\epsilon > 0$, there exist an integer $m$, an $n \x m$ matrix $\bfA$, an $m \x n$ matrix $\bfB$, and an $m$-dimensional vector $\bm{\theta}$ such that
\begin{equation}
\max_{\bfq \in K} \| F(\bfq) - \bfA\bm{\sigma}(\bfB\bfq + \bm{\theta})  \| < \epsilon
\end{equation}
holds where $\bm{\sigma}: {}^m\mathbb{R} \rightarrow {}^m\mathbb{R}$ is a sigmoid mapping defined by $\bm{\sigma}({\mbf x}) = \text{col}\,[\sigma(x_i)]$
\label{thm2}
\end{thm}

Taken together, Theorem~\ref{thm1} and Theorem~\ref{thm2} demonstrate that a recurrent neural network~\ref{eq:G} can always be constructed to approximate, arbitrarily well, a given dynamical system~\ref{eq:fn1}. The problem becomes one of constructing a three-layer feedforward neural network that approximates the dynamical system's vector field sufficiently closely.

Before moving on, we note that the time constant $\tau$ must be selected to satisfy certain conditions for Theorem~\ref{thm1} to hold. These are \citep{funahashi1993}
\begin{equation}
\left\|\frac{{\mbf q}}{\tau}\right\| < \frac{\epsilon L_F}{4(e^{L_F T}-1)},\qquad \left\|\frac{1}{\tau}\right\| < \frac{L_G}{2},\qquad \left\|\frac{\bm{\theta}}{\tau}\right\| < \frac{\epsilon L_G}{4(e^{L_G T}-1)},
\label{eq:tau}
\end{equation}
where $L_{G}$ is the Lipschitz constant of $G$. Clearly, $\tau$ satisfies these when it is sufficiently large.

\subsection{Network training and solution}
\label{sub:train}

In light of the foregoing, our aim is to train the three-layer feedforward neural network given by~\ref{eq:ffn} to replicate the vector field $F(\bfq)$. Indeed, 
$E_{\max}$, which we can interpret as the (maximum) training error for the feedforward network, establishes an upper bound on the accuracy of the RNN: Assuming $\tau$ is sufficiently large,
\begin{equation}
    \max_{t \in I} \|\bfq(t) - \bfom(t) \| \leq \frac{2E_{\max}}{L_F} (e^{L_FT} - 1) \triangleq E_L.
\label{eq:app}
\end{equation}
This inequality bounds the error of the RNN trajectory through time by the error of the feedforward network $F_\text{FF}(\bfq)$ over its domain.

Of course, the network $F_\text{FF}(\bfq)$ is amenable to standard machine learning techniques. Facilitating a training approach is the mathematical expression for the vector field $F({\mbf q})$ defining the desired system. This expression can be sampled to arbitrary resolution to provide training data. Specifically, a set of training points can be chosen from some region of interest in the dynamical system's state space. These points then act as the feedforward network input; evaluating $F({\mbf q})$ on the set of input points yields the target set of feedforward outputs. A note about this region of interest from which we sample training data: Recalling the statement of Theorem~\ref{thm1}, the dynamical trajectory $\bfq(t)$ that we wish to approximate must be included in the compact subset $\tilde{K} \subset {}^n\mathbb{R}$ for all $t \in I$. There are two practical consequences of this detail. First, the systems we approximate by RNN should lie in a compact subset of ${}^n\mathbb{R}$ for all time, in order for our approximations to hold. As we are primarily interested in systems with attractors, this requirement will be satisfied. Second, the compact subset $\tilde{K}$ provides a natural envelope for the sampling domain of our training data.

We have implemented a machine-learning approach with excellent results. Using stochastic gradient descent, we train a three-layer feedforward neural network on data sampled from vector field $F({\mbf q})$.  Training yields ${\mbf A}$, ${\mbf B}$, and $\bm{\theta}$ such that $E_\text{max}$ is small. These feedforward networks are then transformed into continuous-time recurrent networks as described in the previous subsection. Specifically, the recurrent network's ODE and initial condition are given by~\ref{eq:G}. The recurrent network so defined closely reproduces the original system's dynamics, as results presented in Section~\ref{sec:res} demonstrate. In practice, we find we can do much better than the bound in \ref{eq:app}, which increases exponentially with time.

\subsection{Driven Systems}
\label{sub:force}

We have thus far concentrated entirely on the simulation of autonomous dynamical systems (and so it was with~\cite{funahashi1993}). However, another important class of problem to consider is the system with external forcing.

Driven systems are generally described by nonautonomous ODEs, namely, in the case of an additive driving signal,
\begin{equation}
	\dot{{\mbf q}} = F({\mbf q}) + {\mbf f}(t),
	\label{eq:driven}
\end{equation}
where ${\mbf f}(t)$ accounts for any external force. It is certainly possible to drive a network constructed with our algorithm by some external signal ${\mbf f}(t)$. This is just a matter of connecting the driving signal additively to the system's output units and integrating as usual (when the driving signal is additive as in~\ref{eq:driven}). However, we want to show that it is possible for our framework to capture driven systems in their entirety; that is, that a properly trained and constructed recurrent network can generate the solution of the complete system~\ref{eq:driven}.

As discussed in the introductory section, our motivation in taking this approach is the decomposition of dynamical systems and the combination of separately trained RNNs. Decomposition is interesting in its own right from the reductionist perspective, but we have also found in our experiments that decomposition is practically useful in training RNNs on complicated dynamical systems.

Driven systems do not fall under the realm of Funahashi and Nakamura's theoretical work, but \cite{chow2000} made this extension. They specifically treat the nonautonomous dynamical system $\dot{\bfq} = F(\bfq, {\mbf u}, t)$ where ${\mbf u} \in {}^{n_\text{in}}\mathbb{R}$ is an autonomous input signal and $t$ is time. It is proved that, for any bounded input ${\mbf u}: [0, \infty) \rightarrow D_U$, a compact set, an RNN exists whose output trajectories stay arbitrarily close to the orbits of the nonautonomous system. This RNN takes the form
\begin{equation}
\dot{\mbf s} = -\frac{1}{\tau}{\mbf s} + {\mbf W}_1 \bm{\sigma}({\mbf s} + {\mbf W}_2 {\mbf u} + {\mbf W}_3 t),
\end{equation}
where ${\mbf W}_1$ is as ${\mbf W}$ in~\ref{eq:W},
\begin{equation}
    {\mbf W}_2
	   = \left( \begin{array}{c}
		  \Zero \\
		  {\mbf B}_2 \end{array} \right) \in {}^{n+m}\mathbb{R}^{n_\text{in}},
\label{eq:W2}
\end{equation}
and ${\mbf W}_3 \in {}^{n+m}\mathbb{R}$ is a vector. 

The alternative approach that we take is to recognize that certain forcing functions can themselves be considered products of their own dynamics. For example, a sinusoidal forcing term, $f(t) = a \cos (\omega t + \phi)$, is the result of the system $\ddot{f} + \omega^2 f = 0$, where the amplitude and phase are determined by the initial conditions. Thus, it is possible to build a recurrent neural network from the forcing dynamics in the same manner as described previously for autonomous systems. This forcing network can be combined with a network simulating the unforced dynamics to yield a {\it forced recurrent neural network} (FRNN).

Technically, construction of the FRNN involves rewriting the $n$-dimensional system~\ref{eq:driven} as an augmented $(n+p)$-dimensional system, $\xdot = R(\bfx) + \bfE\bfx$, where the function $R$ accounts for the dynamics of $F$ {\it as well as} the dynamics of the driving system itself. Dynamics of the driving system are represented in the $p$ additional state variables. The matrix $\bfE$ describes how the $p$ state variables of the driving system affect (additively) the $n$ state variables of $F$. The following theorem, which we prove in the appendix, guarantees that such an augmented system can be approximated to arbitrary accuracy (in the sense of Theorem~\ref{thm1}) by a RNN.

\begin{thm}
Let $D$ be an open subset of$~{}^{n+p}\mathbb{R}$, $R: D \rightarrow {}^{n+p}\mathbb{R}$ a $C^1$-mapping, $\bfE \in {}^{n+p}\mathbb{R}^{n+p}$ a matrix, and $\tilde{K}$ a compact subset of $D$. Suppose that there is a subset $K \subset \tilde{K}$ such that any solution $\bfx(t)$ with initial value $\bfx(0)$ in $K$ of an ordinary differential equation
\begin{equation}
	\xdot = R(\bfx) + \bfE\bfx
\label{eq:a1}
\end{equation}
is defined on $I = [0, T]$, $0 < T < \infty$ and $\bfx(t)$ is included in $\tilde{K}$ for any $t \in I$.

Then, for arbitrary $\epsilon > 0$, there exist integers $m$ and $q$, and a recurrent neural network with $n+p$ output units and $m+q$ hidden units such that for a solution $\bfx(t)$ satisfying~\ref{eq:a1} and an appropriate initial state of the network,
\begin{equation}
	\max_{t \in I} \| \bfq(t) - \bfom(t) \| < \epsilon
\label{eq:a2}
\end{equation}
holds, where $\bfq \in {}^n\mathbb{R}$ represents the $n$ driven state variables of $\bfx$, and $\bfom \in {}^n\mathbb{R}$ represents an appropriate subset of the RNN's output units.
\label{thm3}
\end{thm}

The required recurrent network, the FRNN, has $n+p$ output units, whose states replicate the $(n+p)$-dimensional orbits of the augmented system~\ref{eq:a1}, and $m+q$ hidden units. The FRNN's particular form is
\begin{equation}
    \dot{\mbf s}_\Sigma = -\frac{1}{\tau}{\mbf s}_\Sigma 
        + {\mbf W}_\Sigma{\bm\sigma}({\mbf s}_\Sigma) + \bfK_\Sigma{\mbf s}_\Sigma.
\label{eq:FRNN}
\end{equation}
We define the terms of~\ref{eq:FRNN} below, as we demonstrate how the FRNN is constructed.

Begin by assuming the dynamics of the driving force to be described by
\begin{equation}
    \qdot_\f = F_\f(\bfq_\f)
\end{equation}
where $\bfq_\f(t) \in {}^p\Re$ and $p \geq n$, the dimension of the original dynamical system. The first $n$ components of $\bfq_\f$ are taken to be ${\mbf f}(t)$, i.e., $\bfq_\f = \text{col}\,[{\mbf f}, {\mbf g}]$.  (In the case, for example, of a forcing function with oscillatory dynamics, this generalized form is necessary.)  Using feedforward training, we teach an RNN the forcing dynamics, representing it as
\begin{equation*}
    \dot{{\mbf s}}_f = G_f({\mbf s}_f)
\end{equation*}
where, in like fashion to $G({\mbf s})$,
\begin{equation}
    G_f({\mbf s}_f) = -\frac{1}{\tau}{\mbf s}_f
        + {\mbf W}_f\bm{\sigma}({\mbf s}_f),
\end{equation}
The state vector ${\mbf s}_f \in {}^{p+q}\mathbb{R}$ includes the $p$ output neurons (${\bm\omega}_f$), which include the forcing functions, and $q$ hidden neurons (${\bm\eta}_f$).  The weight matrix, in familiar manner, is described as
\begin{equation}
	{\mbf W}_f
	   = \left( \begin{array}{cc}
		{\mbf O} & {\mbf A}_f \\
		{\mbf O} & {\mbf C}_f \end{array} \right),
\end{equation}
where ${\mbf C}_f \triangleq {\mbf B}_f{\mbf A}_f$, ${\mbf A}_f \in {}^p\mathbb{R}^q$, and ${\mbf B}_f \in {}^q\mathbb{R}^p$.  We further break down ${\mbf A}_f$ as
\begin{equation}
    {\mbf A}_f \triangleq \left( \begin{array}{cc}
        {\mbf A}_\ff & {\mbf A}_{fg} \end{array} \right)
\end{equation}
where ${\mbf A}_\ff \in {}^n\Re^q$ is associated with the force vector ${\mbf f}$.

As before, a recurrent network, $\dot{{\mbf s}} = G({\mbf s})$, is trained on the vector field of the prescribed dynamical system, $F$.  This training is performed on the system \textit{in the absence} of the driving force. The task is now to merge the two networks, $G$ and $G_{f}$.

We note that the input of the $n$ output neurons $\bfom$ of $G$ is the approximated vector field $\tilde{F}(\bfom)$.  We therefore wish to add as inputs to these neurons, though bypassing the sigmoid operator, the outputs $\bfom_{f}(t)$ from $G_{f}$.  Mathematically, both $G$ and $G_{f}$ may be parsed as
\begin{equation}
\begin{alignedat}{2}
    \dot{\bfom} & = -\frac{1}{\tau}\bfom + \bfA{\bm\sigma}(\bfet), \qquad 
        & \dot{\bfom}_{f} & = -\frac{1}{\tau}\bfom_{f}
        + \bfA_{f}{\bm\sigma}(\bfet_{f}) \\
    \dot{\bfet} & = -\frac{1}{\tau}\bfet + \bfC{\bm\sigma}(\bfet) 
        + \frac{1}{\tau}\bm{\theta}, \qquad
        & \dot{\bfet}_{f} & = -\frac{1}{\tau}\bfet_{f} + \bfC_{f}{\bm\sigma}(\bfet_{f}) + \frac{1}{\tau}\bm\theta_{f}
\end{alignedat}
\end{equation}
We need to augment the differential equation for $\bfom$ to
\begin{equation}
    \dot{\bfom} = -\frac{1}{\tau}\bfom + \bfA{\bm\sigma}(\bfet) + \bfom_\ff.
\label{f:1}
\end{equation}
The new merged system $G_\Sigma$ may then be written as in~\ref{eq:FRNN}, with
\begin{equation*}
    {\mbf s}_\Sigma \triangleq \left(\begin{array}{c} 
        \bfom \\ \bfom_\f \\ \bfet \\ \bfet_\f
        \end{array} \right), \qquad
    {\mbf W}_\Sigma \triangleq \left(\begin{array}{cccc}
        \cdot & \cdot & {\mbf A} & \cdot \\
        \cdot & \cdot & \cdot & {\mbf A}_\f \\
        \cdot & \cdot & {\mbf C} & \cdot \\
        \cdot & \cdot & \cdot & {\mbf C}_\f 
        \end{array} \right), \qquad
    \bfK_\Sigma \triangleq \left( \begin{array}{cccc}
        \cdot & {\mbf K}_{f} & \cdot & \cdot \\
        \cdot & \cdot & \cdot & \cdot \\
        \cdot & \cdot & \cdot & \cdot \\
        \cdot & \cdot & \cdot & \cdot
        \end{array} \right),
\end{equation*}
where $\bfK_f \triangleq [\begin{array}{cc} {\mbf 1} & \Zero \end{array}]$ and ${\mbf 1}$ is the $n\x n$ identity matrix which has the effect of picking out only the variables corresponding to ${\mbf f}$, i.e., $\bfom_\ff$, in $\bfom_\f$; the dots represent zero matrices.  To complete the dynamical description, we take
\begin{equation*}
    {\mbf s}_\Sigma(0) \triangleq \left(\begin{array}{c}
        \bfq(0) \\ \bfq_\f(0) \\ {\mbf B}\bfq(0) + \bm\theta \\ {\mbf B}_\f\bfq_\f(0) + \bm\theta_\f
        \end{array} \right)
\end{equation*}
as the initial conditions, where the forcing bias ${\bm\theta}_{f} \in {}^q\mathbb{R}$. This completes the description of the FRNN.

By Theorem~\ref{thm3}, the solution $\bfom$ computed from the output neurons of the recurrent neural network $G_\Sigma$ is guaranteed to be arbitrarily close to the solution to the true dynamical system $\qdot = F(\bfq) + {\mbf f}(t)$.

\section{Results: dynamical systems realized as RNNs}
\label{sec:res}

We now construct and present a series of recurrent neural systems. Our focus is on systems with attractors in their state space because of their importance to neuroscience and because of the earlier discussion on the set $\tilde{K}$. Recurrent networks are constructed as detailed in \S\ref{sec:rnn}. For training, we use stochastic gradient descent with the {\it Adam} optimizer in the standard configuration~\citep{kingma2014}. Training data is given by samples from a dynamical system's vector field $F({\mbf q})$ on some compact set $\tilde{K}$. Although the number of output units $n$ is defined by the dimensionality of $F$, the number of hidden units $m$ is unconstrained and a matter of choice. More complicated dynamical systems naturally require a larger $m$ and more training data; in practice, we aim for the smallest $m$ such that training converges. The time constant $\tau$ is set in all networks to $10^6$ (variables are nondimensionalized), although we discuss the time constant's effects further on.

For ease of presentation, most of the systems depicted will lie in a two-dimensional state space; however, the algorithm applies without modification to higher dimensions.

The figures that follow will in general depict: (a) a vector field (black arrows) within some compact region $\tilde{K}$; (b) orbits ${\mbf q}(t)$ obtained by integrating the system's mathematical expression from various initial conditions (green lines); (c) orbits $\bfom(t)$ starting at the same initial conditions, obtained by integrating a trained recurrent neural network (blue lines). Figure captions provide training parameters and error measures. These include $d_{\#}$, the number of training datapoints used; $\text{MSE}$, the mean squared error achieved by the trained feedforward network on the vector field $F({\mbf q})$; $E_\text{max}$, the maximum error on the vector field, which corresponds to the quantity in \ref{eq:vferror}; $E_\text{orb}$, the maximum error between points from the original system orbits and the RNN orbits, divided by the time series length in seconds; and $r$, the spectral radius of $\bfW$. We summarize these in Table~\ref{datatable}. The reader will note that the spectral radii are in some cases quite large. This is because the matrix $\bfA$ must take a value that has been squashed by the sigmoid activation function and expand it to the size of values in the original system.

\begin{table}
	\centering
    \begin{tabular}{ | c | l |}
    \hline
    Symbol & Description \\ \hline
    $m$ & Number of hidden units \\ \hline
    $r$ & Spectral radius of $\bfW$ \\ \hline
    MSE & Mean squared error (training set)  \\ \hline
    $E_\text{max}$ & Maximum training error \\ \hline
    $E_\text{orb}$ & Maximum trajectory error (normalized by $T$) \\ \hline
    $T$ & Time length of trajectories \\ \hline
    $d_{\#}$ & Number of training datapoints \\
    \hline
    \end{tabular}
    \caption{Symbols used in simulation captions.}
	\label{datatable}
\end{table}

\subsubsection*{Example 1: A fixed point attractor in $\mathbb{R}^2$}

We begin with a simple dynamical system featuring a fixed point attractor. Let us embed one fixed point at $(x_1,y_1)=(p_1,p_2)$. We define the vector field accordingly using straightforward linear functions. We have
\begin{align}
F({\mbf q}) &= F(x,y) = (a(x-p_1),b(y-p_2))
\label{eq:vf1}
\end{align}
where setting $a < 0$, $b < 0$ engenders the prescribed attractor dynamics.

\begin{figure}
  \centering
  \includegraphics[width=3in]{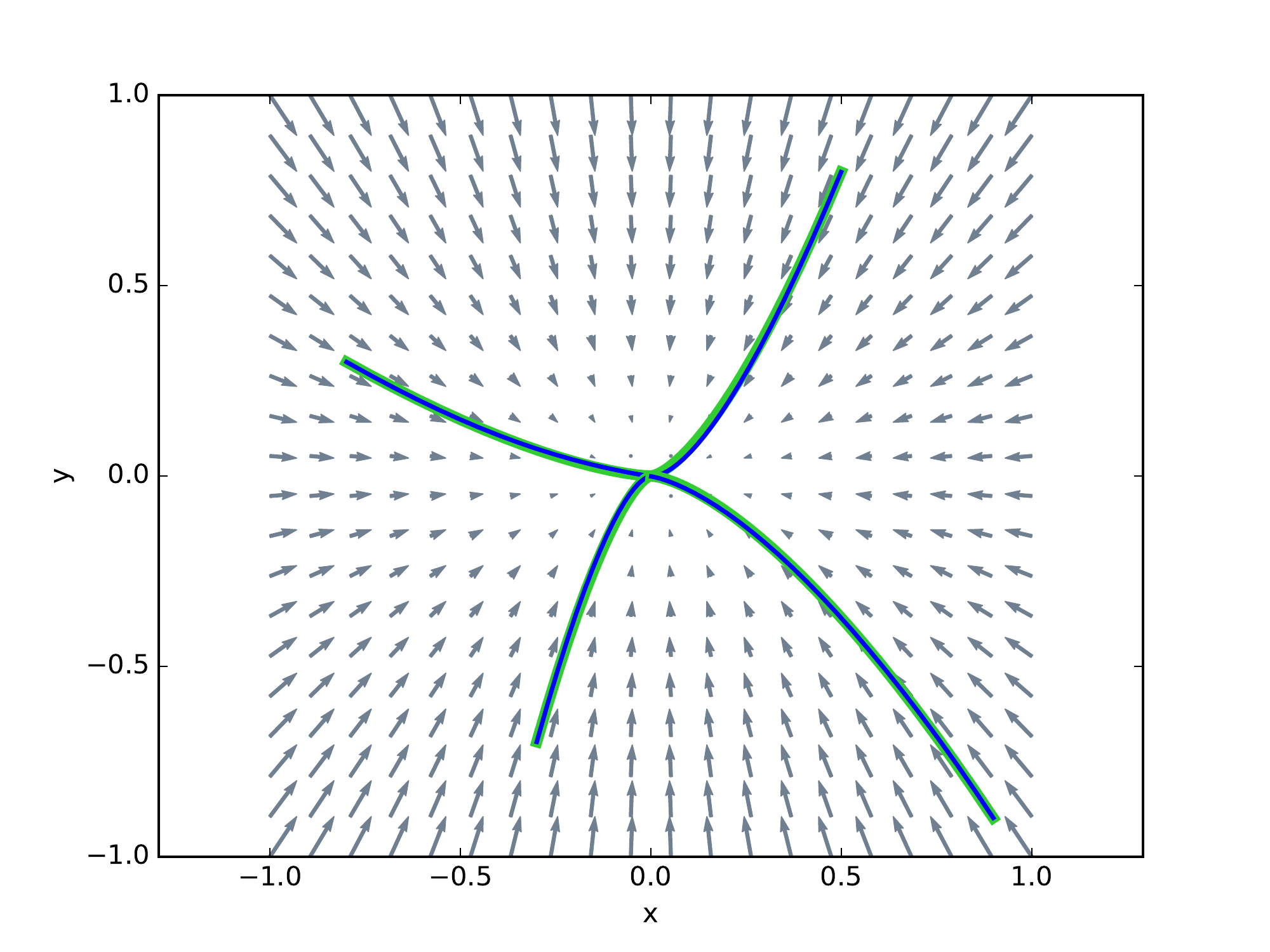}
  \caption{ Vector field and orbits of a system with a fixed point. The original dynamical system is in green, with RNN approximation superimposed in blue. $m=3$, $r=0.302$, $\text{MSE} = 7.5\times 10^{-6}$, $E_\text{max}=8.9\times 10^{-5}$, $E_\text{orb}=3.1\times 10^{-4}$, $T=40$, $d_{\#}=2.5\times 10^{5}$.}
  \label{fig:1fp}
\end{figure}

Figure~\ref{fig:1fp} shows four separate orbits $\bfom(t)$ produced by a RNN trained on the vector field of~\ref{eq:vf1}. These trajectories are superimposed on the vector field itself and orbits ${\mbf q}(t)$ integrated from it. As can be seen, orbits of the recurrent network closely track the trajectories of the dynamical system and converge approximately to the intended fixed point, where they halt. For this simple system only $m = 3$ hidden neurons were required to achieve a very small trajectory error, and the matrices ${\mbf A}$, ${\mbf B}$, and $\bm{\theta}$ after training are
\begin{equation*}
	{\mbf A}
	= 
	\left( \begin{array}{rrrrrrr}
		-1.20327 & -0.07202 & -0.93635 \\
		1.18810 & -1.50015 & 0.93519 \end{array} \right),
\end{equation*}
\begin{equation*}
	{\mbf B}
	= 
	\left( \begin{array}{rr}
		1.21464 & -0.10502 \\
		0.12023 & 0.19387 \\
		-1.36695 & 0.12201 \end{array} \right),
\end{equation*}
\begin{equation*}
	\bm{\theta}
	= 
	\left( \begin{array}{r}
		-7.56499\times 10^{-5} \\
		1.34708\times 10^{-4} \\
		-6.24925\times 10^{-6} \end{array} \right).
\end{equation*}

\subsubsection*{Example 2: Multiple fixed points in $\mathbb{R}^2$}
One of the applications for attractor-based dynamical systems is associative memory~\citep{hopfield1982, bao2012}. In an associative memory, attractor states represent stored patterns that should be recalled, through dynamical evolution, from associated initial conditions.

\begin{figure}
\centering
\subfloat[Vector field and orbits of a system with two fixed points. $m=7$, $r=0.411$, $\text{MSE} = 8.1\times 10^{-6}$, $E_\text{max}=0.0075$, $E_\text{orb}=0.033$, $T=40$, $d_{\#}=1.6\times 10^{7}$.]{%
  \includegraphics[width=2.2in]{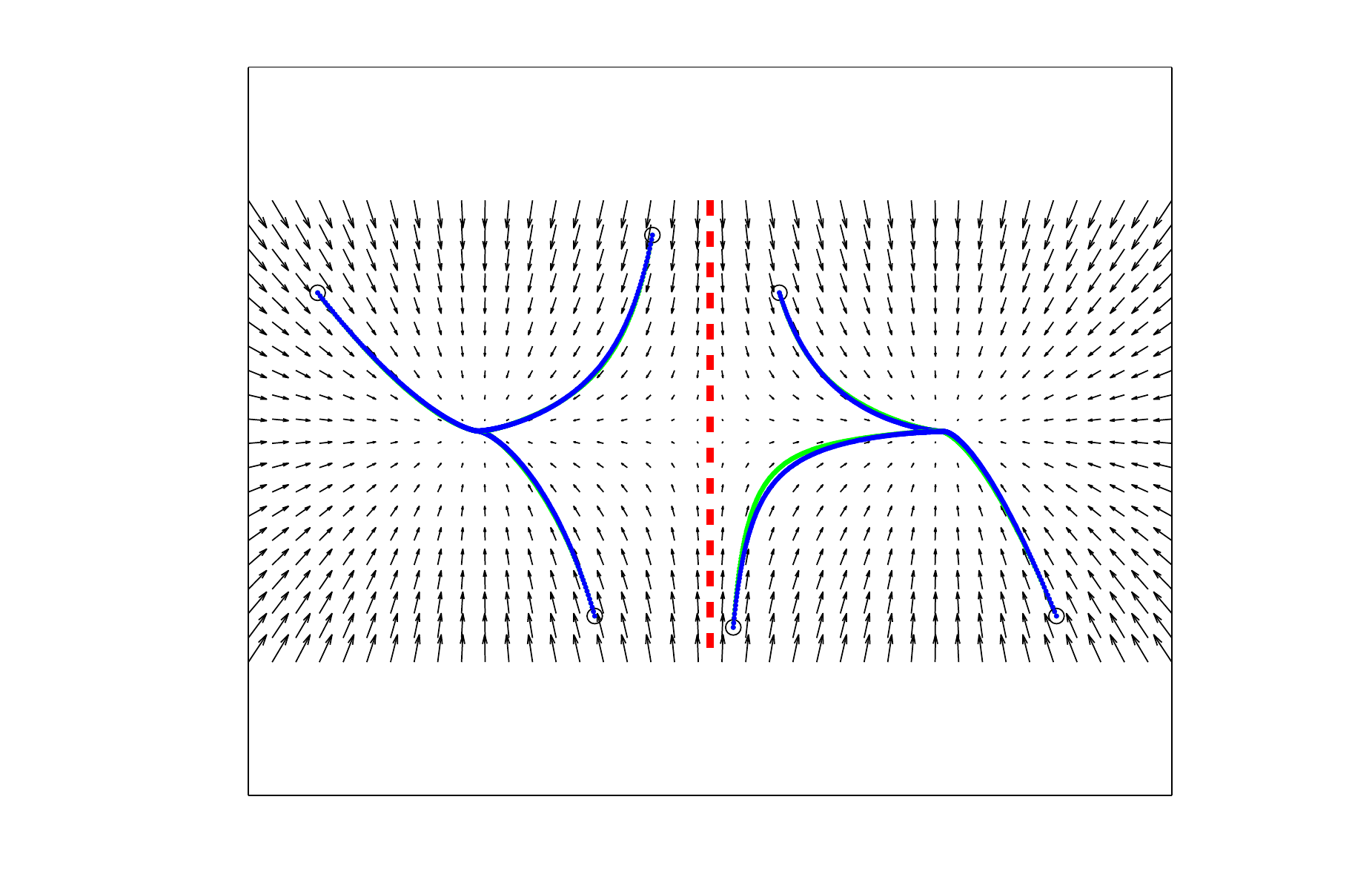}%
}
\qquad
\subfloat[Vector field and orbits of a system with four fixed points. $m=9$, $r=0.492$, $\text{MSE} = 2.2\times 10^{-5}$, $E_\text{max}=0.011$, $E_\text{orb}=0.044$, $T=40$, $d_{\#}=1.6\times 10^{7}$.]{%
  \includegraphics[width=2in]{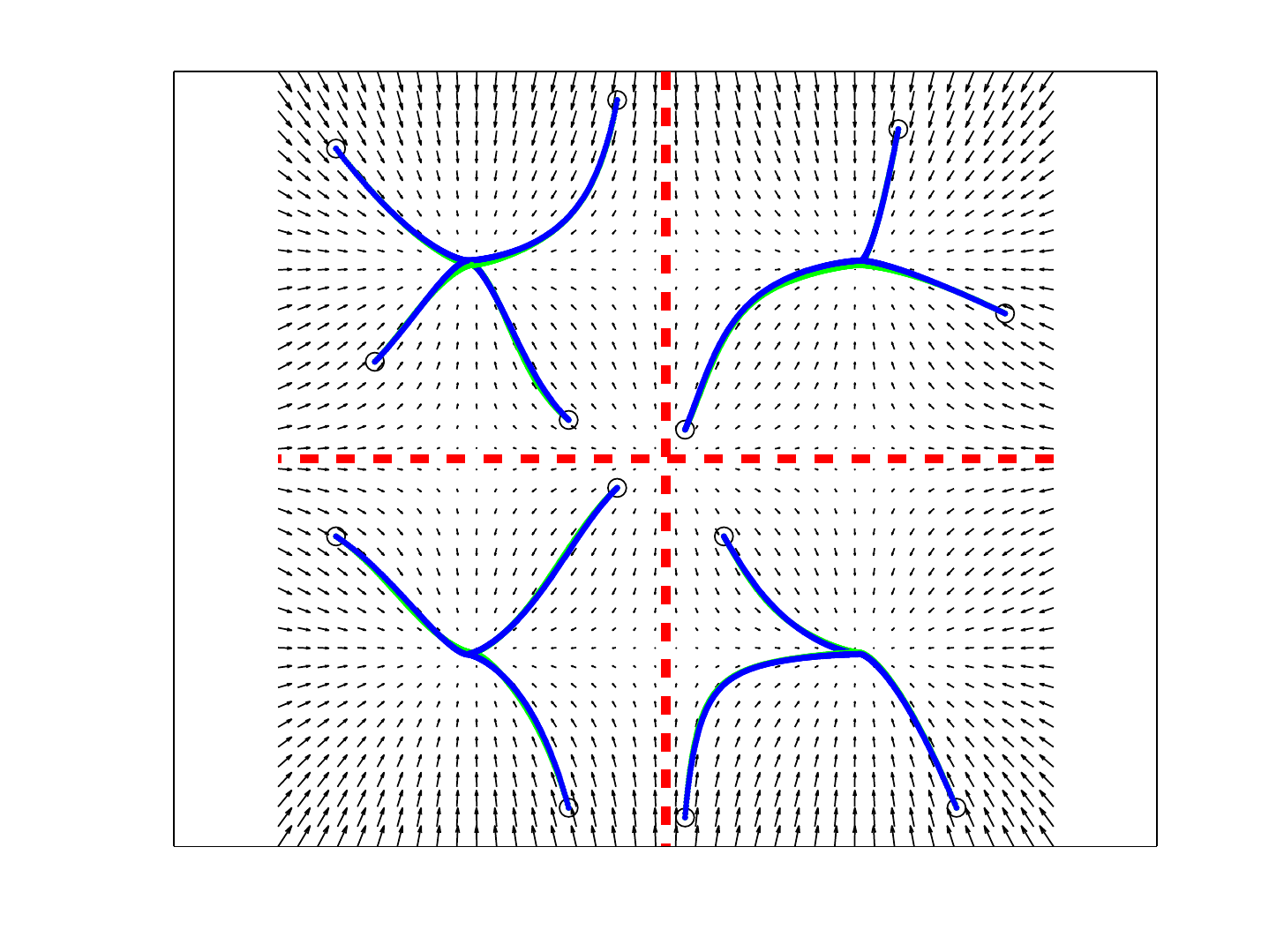}%
}

\caption{Systems with multiple fixed points. The original dynamical systems are in green, with RNN approximations superimposed in blue.}
\label{fig:2fps}
\end{figure}

We can build such a dynamical system using the vector-field regularization technique of~\cite{sotomayor1996}. Regularization is a technique for smoothly combining discontinuous vector fields. For example, the system depicted in Figure~\ref{fig:2fps}(a) contains two fixed points, and is constructed by joining two copies of the vector field from the previous example---one with a fixed point at $(p_1, p_2)$ and the other with a fixed point at $(-p_1, p_2)$ (see~\cite{trischler2013} for a more detailed description of this method). We trained a RNN on this more complicated dynamical system, which features two basins of attraction and a separatrix in between. It required $m=7$ hidden neurons to approximate to similar accuracy to the previous case. We can continue adding fixed points in this manner; as an additional example, we present a regularized system of four fixed points and its approximating RNN trajectories in Figure~\ref{fig:2fps}(b).

\subsubsection*{Example 3: A limit cycle in $\mathbb{R}^2$}
\begin{figure}
  \centering
  \includegraphics[width=3in]{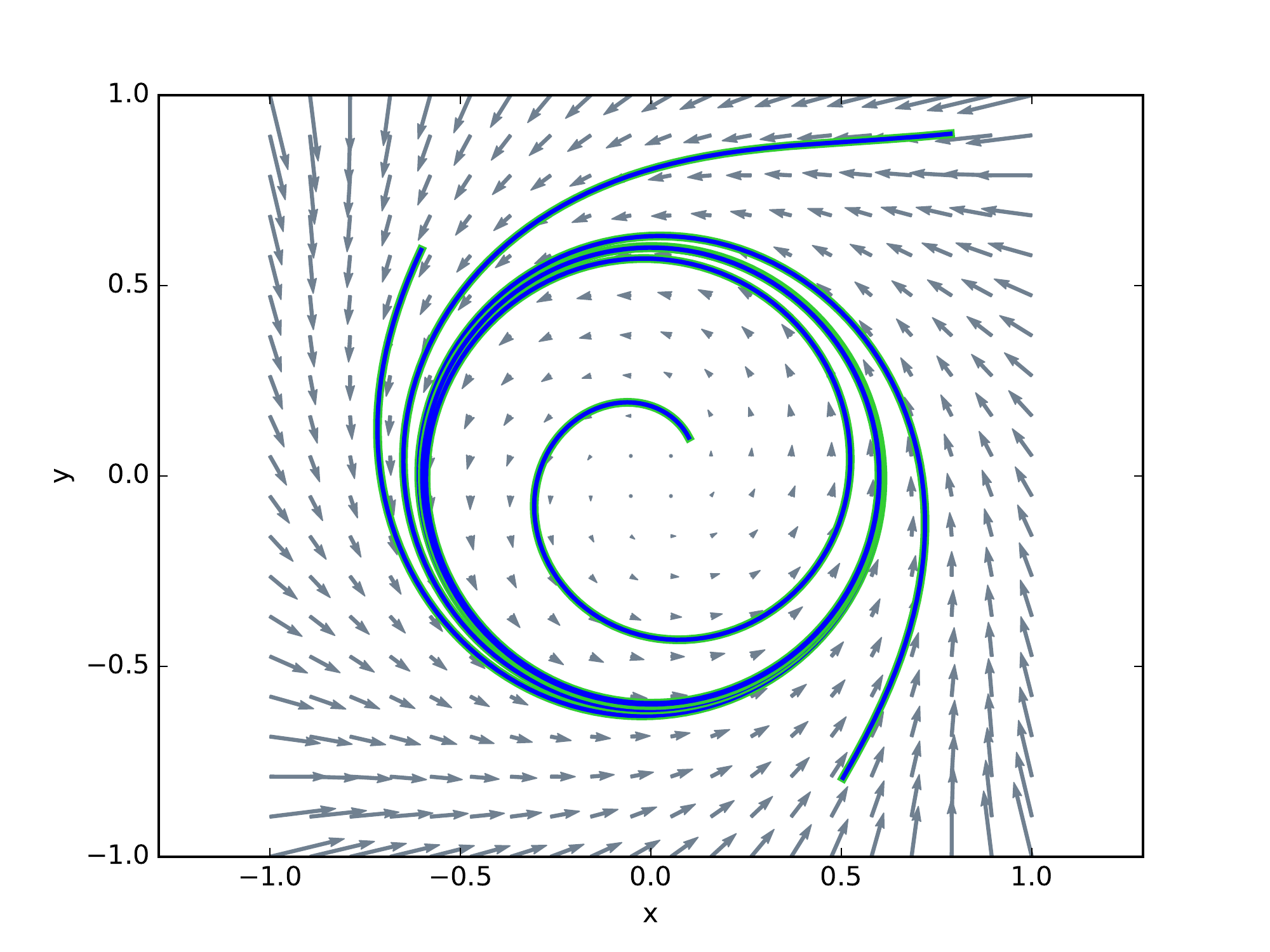}
  \caption{ Vector field and orbits of a system with a limit cycles. The original dynamical system is in green, with RNN approximation superimposed in blue. $m=30$, $r=7.13$, $\text{MSE} = 1.4\times 10^{-5}$, $E_\text{max}=0.0013$, $E_\text{orb}=3.2\times 10^{-4}$, $T=40$, $d_{\#}=1.6\times 10^{7}$.}
  \label{fig:1lcs}
\end{figure}

The 2-dimensional vector field given by
\begin{align*}
\dot{x} &= -y + x(r^2 - x^2 - y^2)\\
\dot{y} &= x + y(r^2 - x^2 - y^2)
\end{align*}
exhibits a globally attracting limit cycle, centered at the origin with radius $r$. In contrast to the preceding example this is a nonlinear system. Stable limit cycles like this are biologically important because they characterize the dynamics of central pattern generators~\citep{hooper2001}. In Figure~\ref{fig:1lcs} we plot trajectories from this system and from an approximating RNN. Low approximation error was achieved with a network of $m=30$ hidden neurons.

\subsubsection*{Example 4: The Van der Pol oscillator}

\begin{figure}
  \centering
  \includegraphics[width=4in]{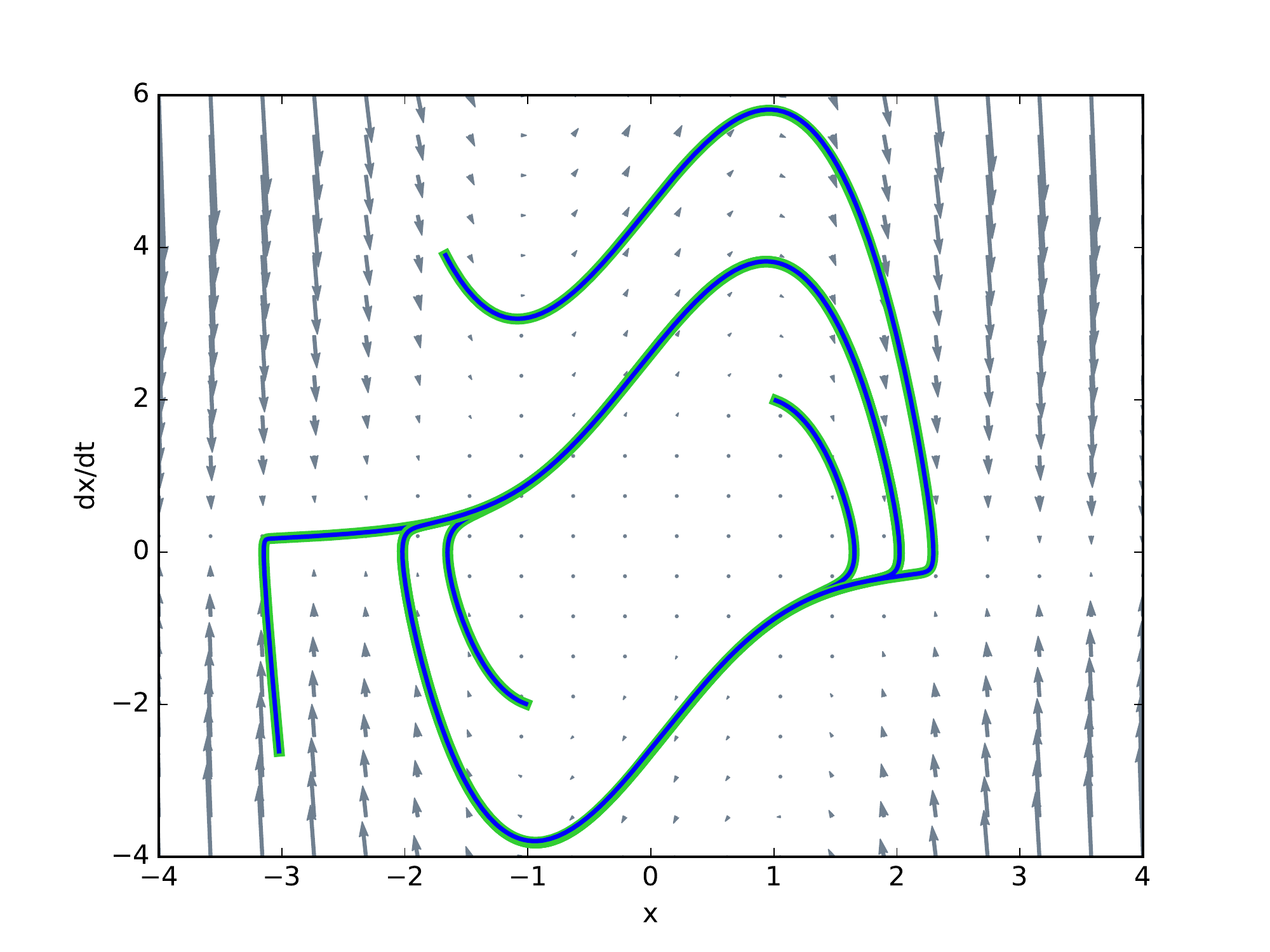}
  \caption{ Vector field and orbits for the Van der Pol system. The original dynamical system is in green, with RNN approximation superimposed in blue. $m=50$, $r=85.2$, $\text{MSE} = 3.4\times 10^{-3}$, $E_\text{max}=0.057$, $E_\text{orb}=0.047$, $T=40$, $d_{\#}=1.6\times 10^{7}$.}
  \label{fig:vdp}
\end{figure}

The well known Van der Pol oscillator is a nonconservative oscillator with nonlinear damping, given by the second-order ODE
\begin{equation}
\ddot{x} + \mu(x^2-1)\dot{x} + \omega^2x = 0.
\label{eq:vdp}
\end{equation}
Parameter $\mu$ gives the damping strength. This system exhibits a limit cycle that becomes increasingly sharpened as $\mu$ increases. Extensions of the model have been used in various fields, most notably to model action potentials in neurons. A two-dimensional phase portrait of this oscillator, along with close RNN approximations of the displayed trajectories, is given in Figure~\ref{fig:vdp}.

\subsubsection*{Example 5: The forced Duffing oscillator}

\begin{figure}
  \centering
  \includegraphics[width=3in]{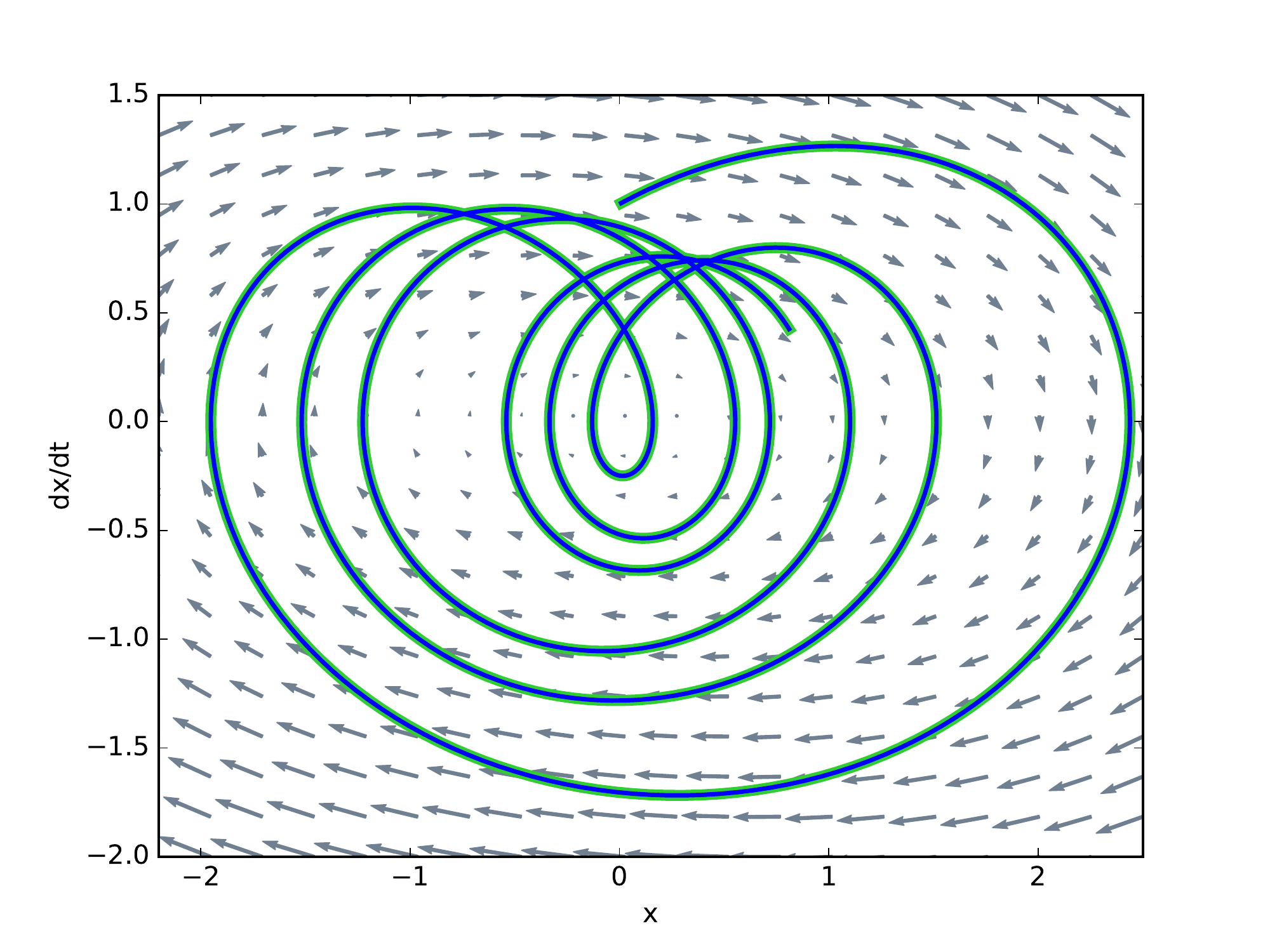}
  \caption{ Vector field and orbits for the forced Duffing equation. The original dynamical system is in green, with RNN approximation superimposed in blue. $m=10+4$, $r=0.303$, $\text{MSE} = 2.3\times 10^{-6}$, $E_\text{max}=5.4\times 10^{-3}$, $E_\text{orb}=7.1\times 10^{-4}$, $T=40$, $d_{\#}=1.6\times 10^7$.}
  \label{fig:duff}
\end{figure}

The Duffing equation describes a damped harmonic oscillator with an additional cubic `hardening' term. When driven by sinusoidal forcing the equation has a highly complex response. The forced Duffing equation is given by the second-order ODE
\begin{equation}
\ddot{x} + 2\zeta\omega\dot{x} + \omega^2(x +\alpha x^3) = f_0 \cos t,
\end{equation}
where $\omega$ is the natural frequency, $\zeta$ is the damping coefficient, $\alpha$ is the nonlinear stiffness parameter, and $f_0$ is the forcing amplitude. A 2-dimensional phase portrait of this oscillator, with parameter settings $\omega=0.5$, $\zeta=0.1$, $\alpha=0.05$, $f_0 = 5/8$, is given in Figure~\ref{fig:duff}. This is an example of a driven dynamical system, and we build our approximation as a FRNN. The forcing RNN contained 4 hidden neurons while the homogeneous RNN contained 10 hidden neurons.

\subsubsection*{Example 6: Chaotic attractors in $\mathbb{R}^3$}

\begin{figure}
  \centering
  \includegraphics[width=4.5in]{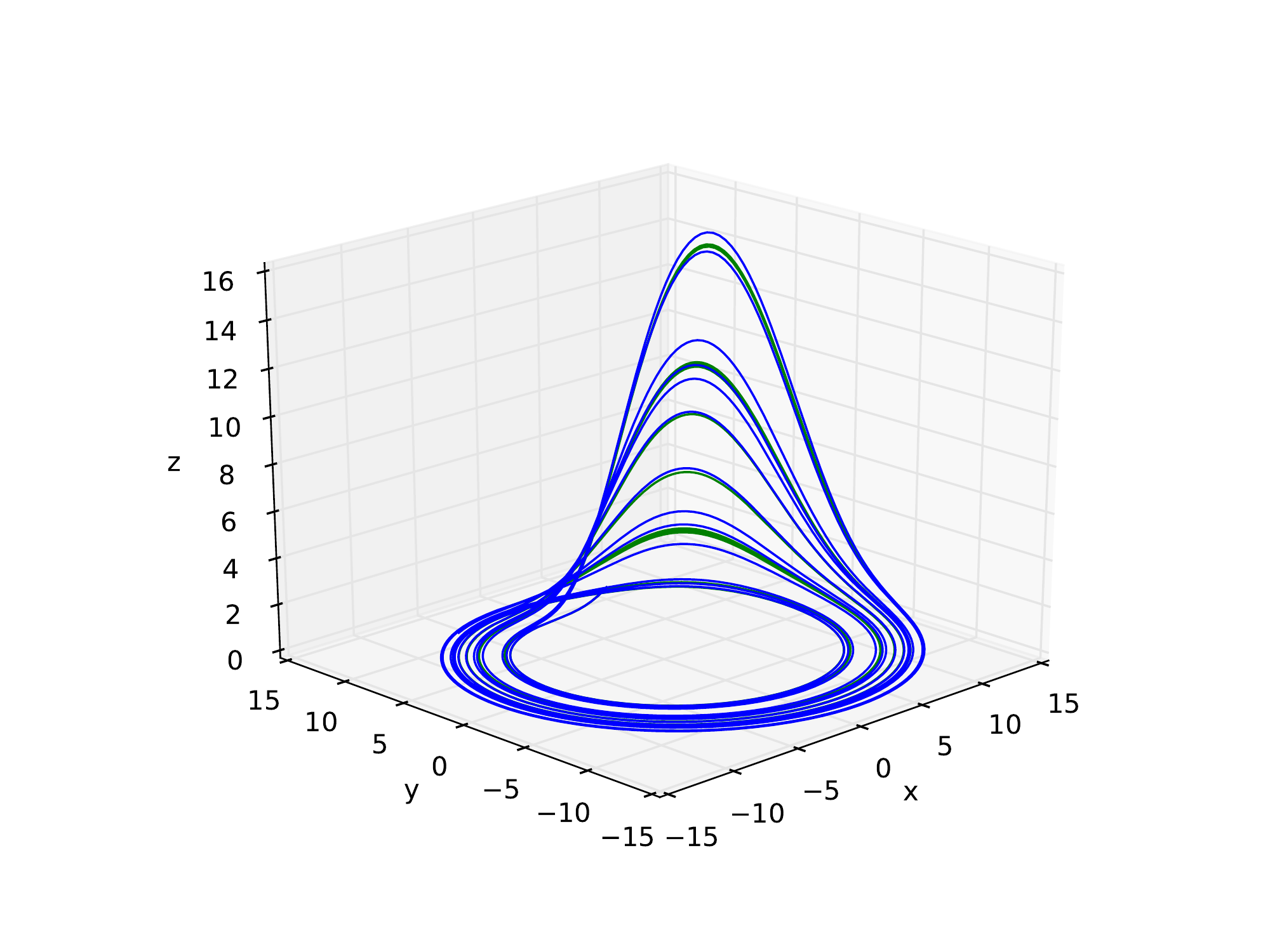}
  \caption{ Orbit of the chaotic R\"{o}ssler system with RNN approximation. The original dynamical system is in green, with RNN approximation superimposed in blue. $m=40$, $r=47.1$, $\text{MSE} = 0.00267$, $E_\text{max}=0.0053$, $E_\text{orb}=1.8\times 10^{-4}$, $T=80$, $d_{\#}=2.7\times 10^{7}$.}
  \label{fig:ross}
\end{figure}

\begin{figure}
  \centering
  \includegraphics[width=3.5in]{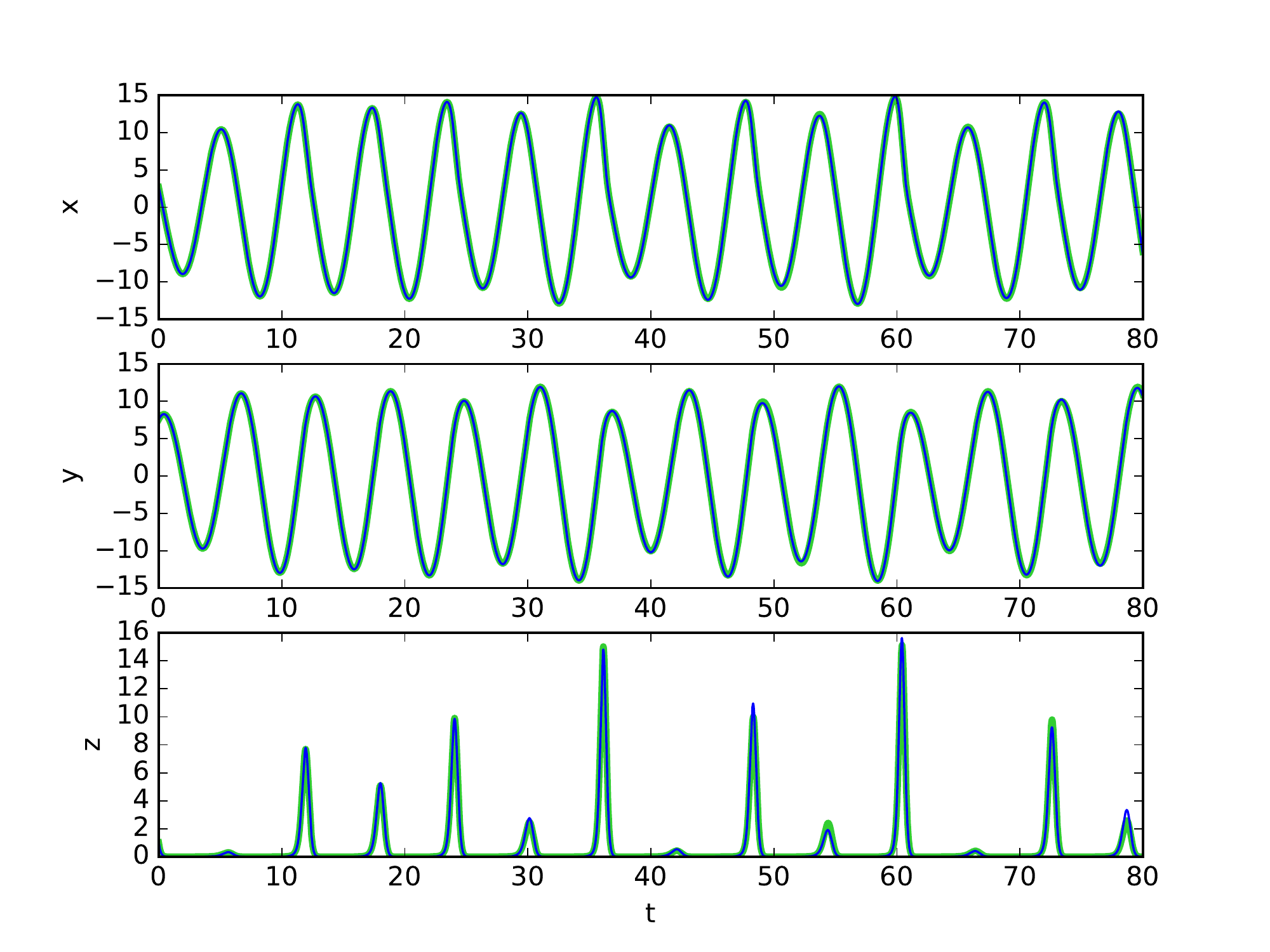}
  \caption{ Time series for the dynamical variables of the chaotic R\"{o}ssler system. RNN approximation in blue. }
  \label{fig:ross_t}
\end{figure}

We now construct a recurrent neural network that replicates the attractor of the R\"{o}ssler system. This three-dimensional system is defined as
\begin{align*}
\dot{x} &= -y - z\\
\dot{y} &= x + ay\\
\dot{z} &= b + z(x-c).
\end{align*}
Setting the system parameters to $a=0.1$, $b=0.1$, $c=9$ yields chaotic dynamics and an attractor with fractal structure, characterized by Cantor-set-like bands and a half-twist as in the M\"{o}bius strip. Despite this complexity, a recurrent network trained on the vector field is able to reproduce the attractor quite closely, as shown in Figure~\ref{fig:ross} (the vector field has been omitted for clarity to better illustrate the structure). Because this plot becomes messy with thicker lines, and it is also difficult to see the green trajectory of the original system, we also present the individual time series for each degree of freedom in Figure~\ref{fig:ross_t}. From this it is clear that the approximation is very good. The depicted RNN contains $m=40$ hidden neurons.


The Lorenz system, developed as a simplified mathematical model for atmospheric convection, is given by
\begin{align*}
\dot{x} &= \sigma(y-x)\\
\dot{y} &= x(\rho - z) - y\\
\dot{z} &= xy - \beta z.
\end{align*}

\begin{figure}
  \centering
  \includegraphics[width=4.5in]{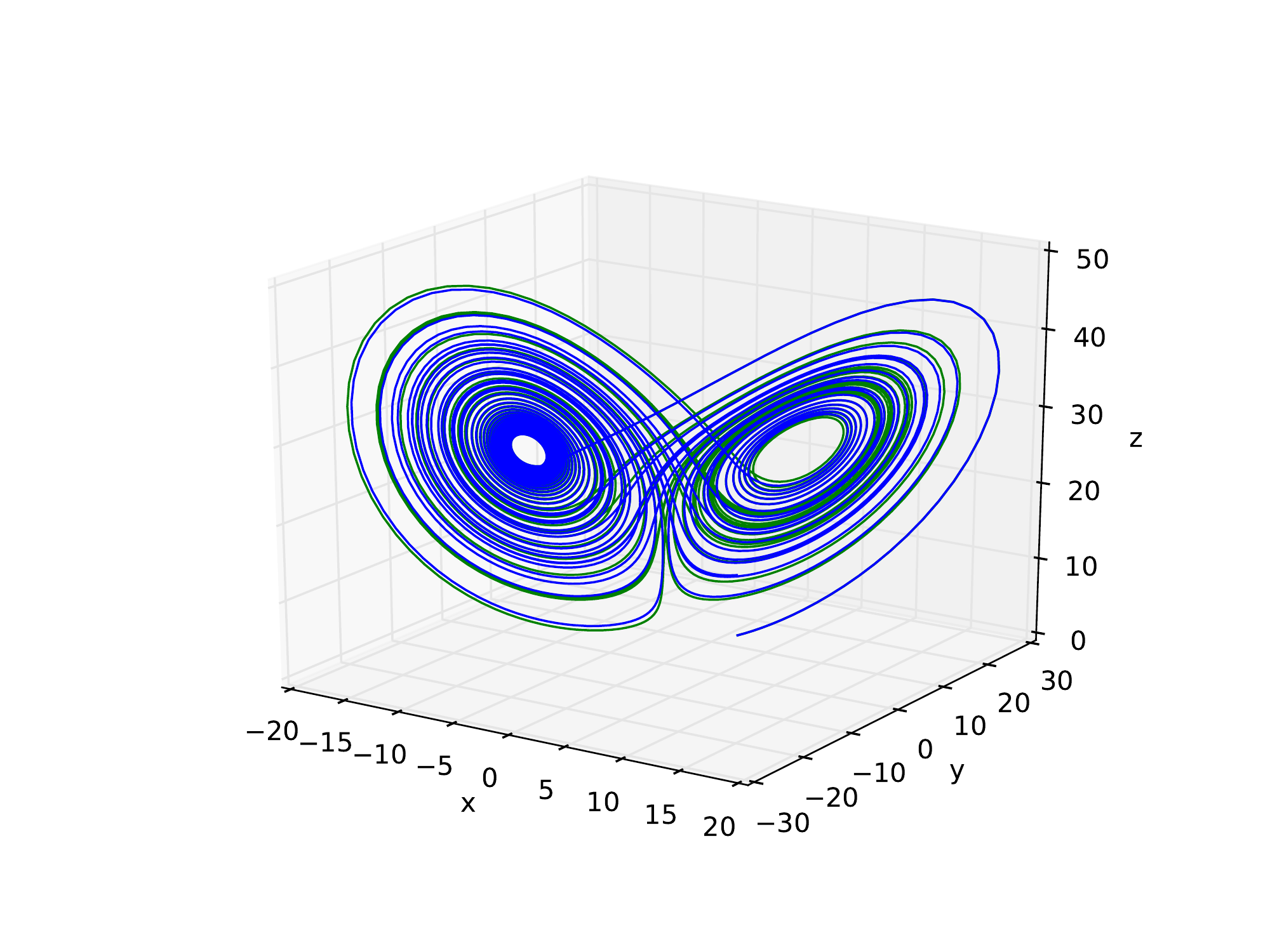}
  \caption{ Orbit of the chaotic Lorenz system with RNN approximation. The original dynamical system is in green, with RNN approximation superimposed in blue. $m=60$, $r=105$, $\text{MSE} = 0.0367$, $E_\text{max}=0.249$, $E_\text{orb}=0.0079$, $T=80$, $d_{\#}=2.7\times 10^{7}$}
  \label{fig:lor}
\end{figure}
\begin{figure}
  \centering
  \includegraphics[width=4in]{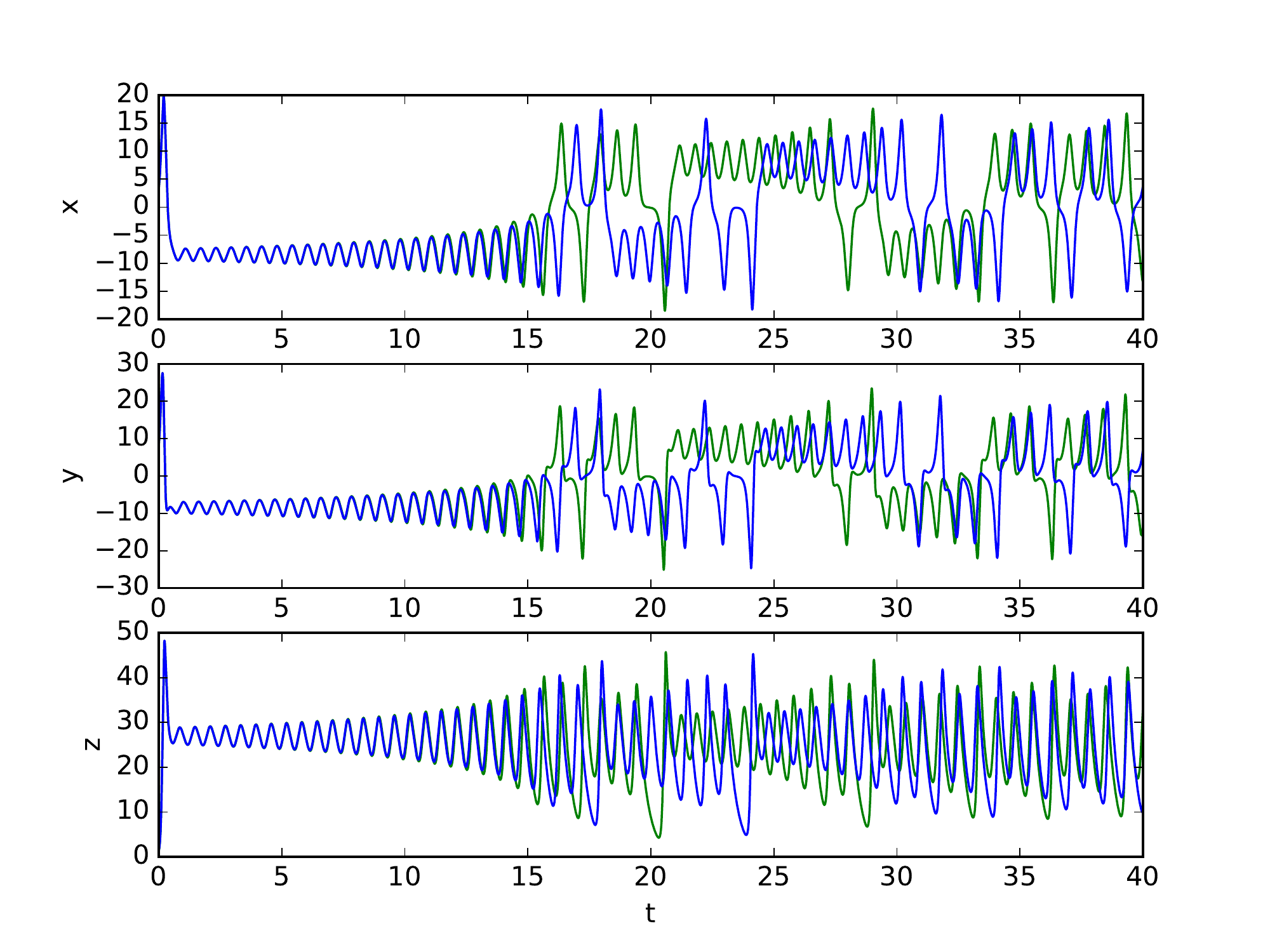}
  \caption{ Time series for the dynamical variables of the chaotic Lorenz system. RNN approximation in blue. }
  \label{fig:lor_t}
\end{figure}
This system exhibits chaotic behaviour for the parameter settings $\sigma = 10$, $\beta = 8/3$, $\rho = 28$, manifest in the butterfly-like attractor presented in Figure~\ref{fig:lor}. The system, having near-periodic orbits circling around two points, is more complicated to learn than the R\"{o}ssler system and required 60 hidden neurons to replicate approximately via RNN. In Figure~\ref{fig:lor_t} we present the time series for each dynamical variable since, as in the R\"{o}ssler case, it is difficult to see the green trajectory of the original system. This figure makes it clear that the RNN captures the quality of the system, but that its timing is off. It looks like the same chaotic system started at a different initial condition. The observed divergence is not a surprise, given the sensitivity of chaotic systems.

\section{Analysis and discussion}
\label{sec:disc}
\subsection{The time constant $\tau$}
The choice of a very large neuronal time constant ($\tau = 10^{6}$) is a safe one, meant to guarantee that the conditions enumerated by~\cite{funahashi1993} are satisfied. We can generally decrease $\tau$ by several orders of magnitude with negligible effect on the accuracy of our RNN approximations. When $\tau$ becomes too small, fixed points are born in the phase space. This is because the time constant governs (inversely) the inhibitory activity of the network; a smaller $\tau$ leads to dissipative dynamics. The effect of decreasing $\tau$ can thus be counteracted by increasing the spectral radius of the weight matrix $\bfW$. In the case of systems where fixed points already exist, these tend to shift location towards the origin, with the shift becoming noticeable for $\tau$ on the order of $10^2$. For systems with limit cycles, the cycles become stable spirals for $\tau$ on the order of $10^4$. Chaotic systems are quite resilient to the formation of fixed points, although the behavior of the systems changes noticeably with decreasing $\tau$. For instance, in the Lorenz system activity tends to concentrate on one of the two wings as $\tau$ gets small, but a fixed point is not born until $\tau$ is on the order of $10^0$.

\begin{figure}[htp]
\centering
\subfloat[$\tau=10^5$]{%
  \includegraphics[width=4in]{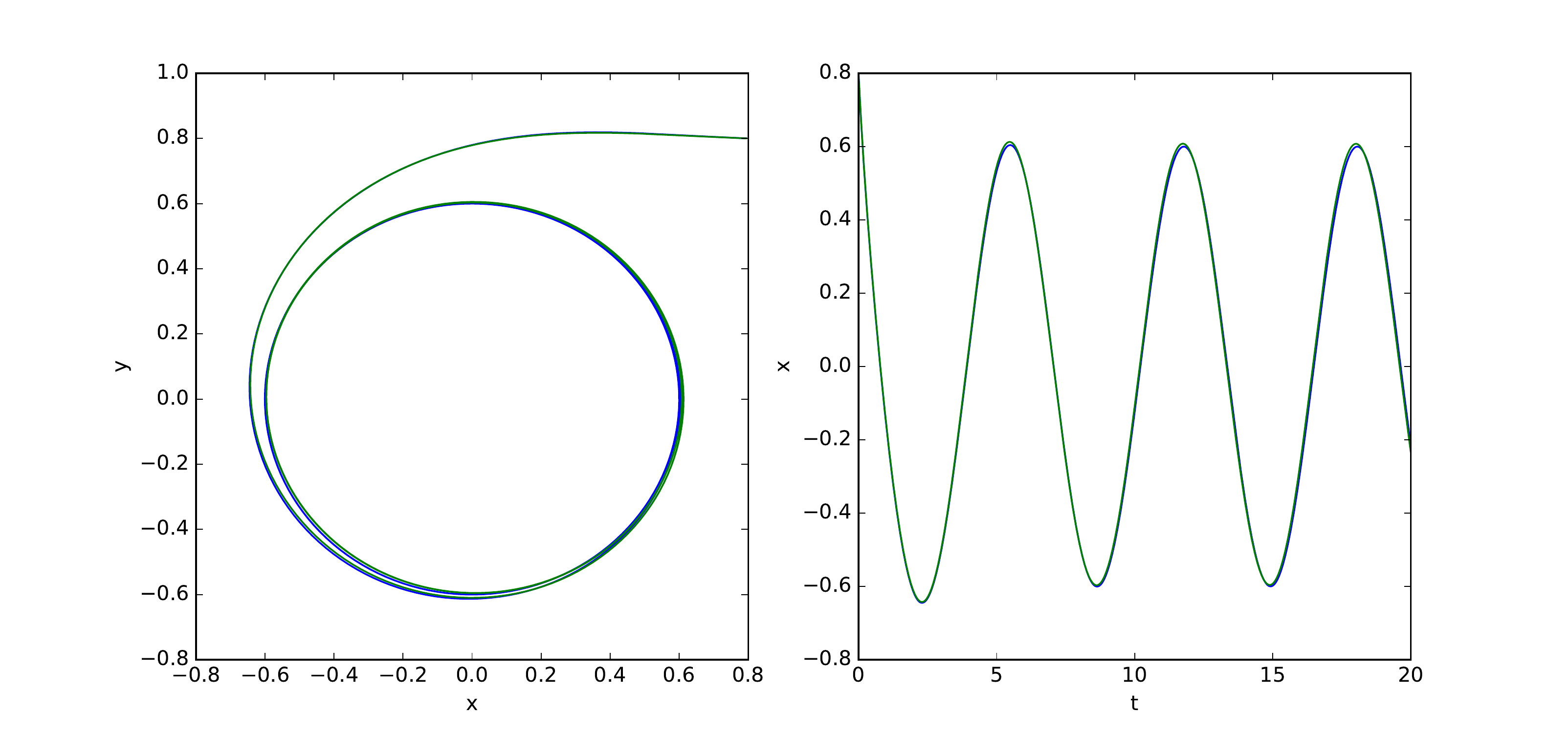}%
}

\subfloat[$\tau=10^3$]{%
  \includegraphics[width=4in]{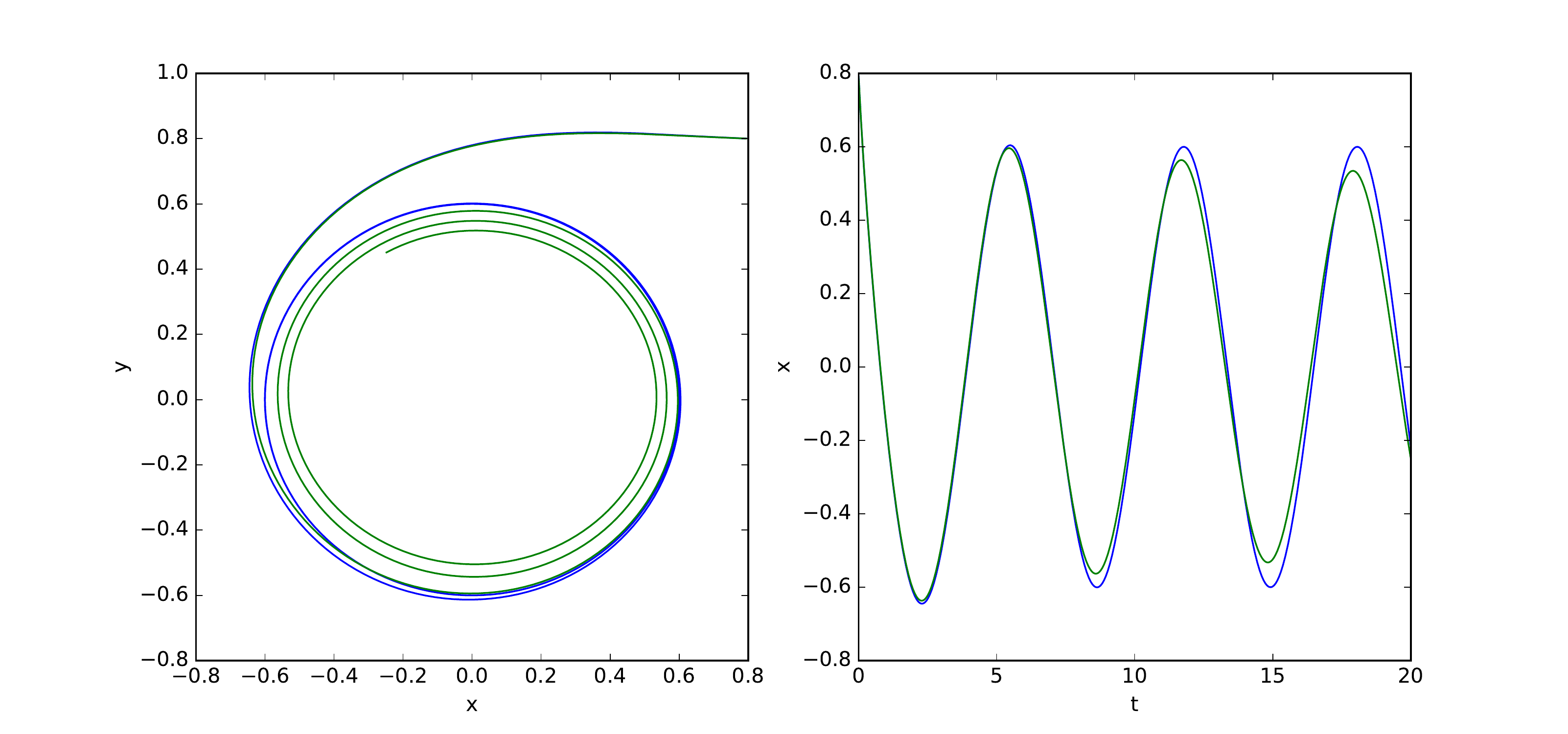}%
}

\subfloat[$\tau=10^2$]{%
  \includegraphics[width=4in]{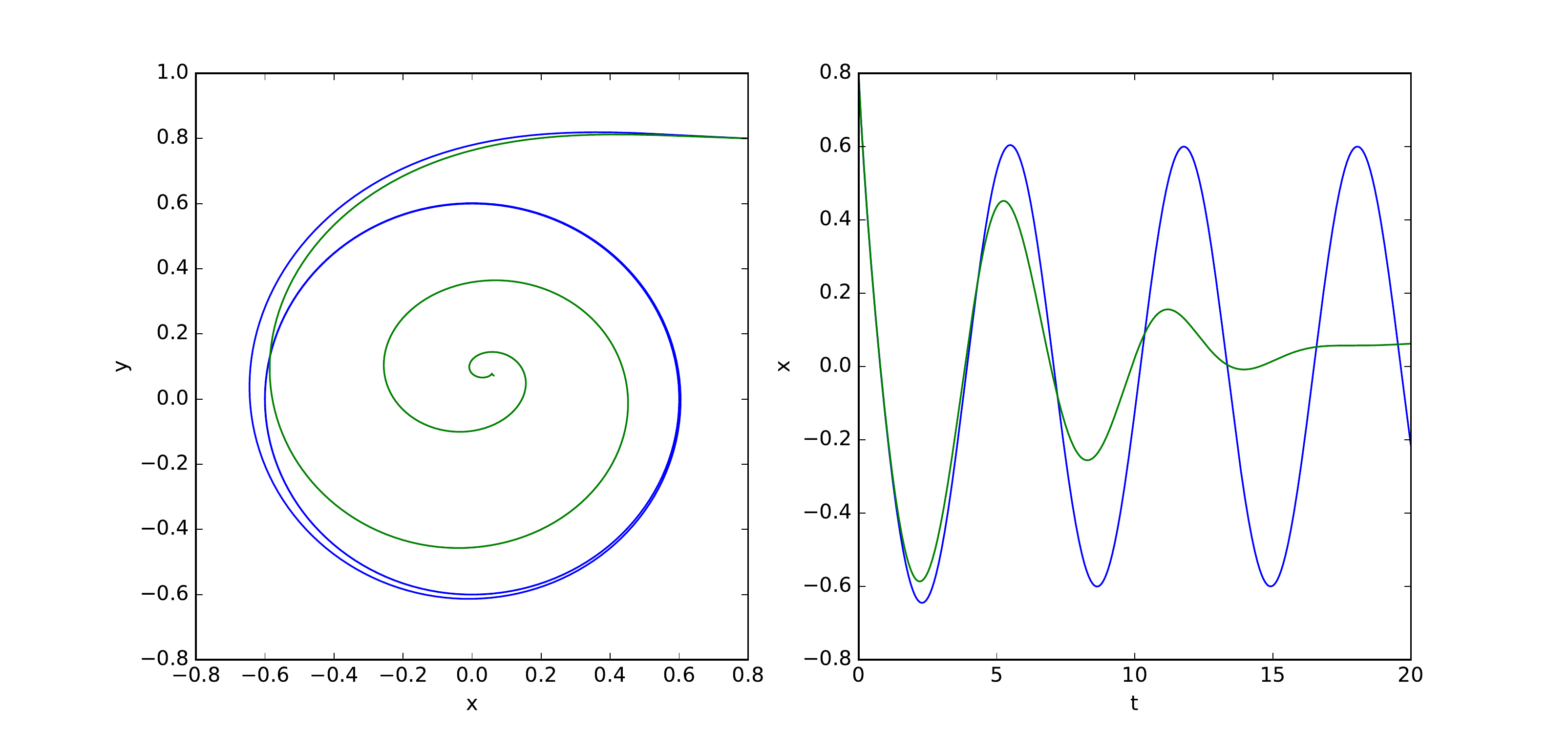}%
}

\caption{Dynamics of the limit-cycle RNN as $\tau$ decreases. The original RNN with $\tau=10^6$ is in blue.}
\label{fig:taus}
\end{figure}

In Figure~\ref{fig:taus} we plot the responses of the limit-cycle RNN in Example 3 as $\tau$ shrinks, to illustrate this behavior. We will also see $\tau$ play an important role in the next subsection.

\subsection{Algebraic approximation of hidden neuron states}
The function of the RNN output neurons is clear: their states replicate the trajectories of the prescribed $n$-dimensional dynamical system being modelled. The hidden neurons are more opaque. We now develop an algebraic approximation to the hidden neurons states, showing that, to within a linearly increasing difference, the hidden states are given by an affine transformation of the output states. The existence of such an approximation makes intuitive sense because the underlying system being modelled is $n$-dimensional.

Note that the hidden neurons are initialized at $t=0$ as $\bfet(0) = {\mbf B}\bfom(0) + \bm{\theta}$. Let ${\mbf r}(t) = {\mbf B}\bfom(t) + \bm{\theta}$. Further, let $\delta(t) = ||\bfet(t) - {\mbf r}(t)||$, and note that $\delta(0) = 0$ exactly. We will show that $\dot{\delta}(t) = \tau^{-1}||\bm{\theta}||$, so that when $\tau$ is large, $\dot{\delta}(t) \ll 1$. Thus, $\delta(t)$ stays near zero for $t$ sufficiently less than $\tau$. Such a statement should be true, else trajectories initialized at different points along a single flow would not match. By this logic, and by the similarity to the form of the initial condition, the validity of approximation does not come as a surprise.

Dropping the explicit time dependence and using the standard RNN equations, we have that
\begin{equation}
\dot{\bfom} = -\frac{1}{\tau}\bfom + {\mbf A}\sigma(\bfet)
\end{equation} and 
\begin{equation}
\dot{\bfet} = -\frac{1}{\tau}\bfet + {\mbf B}{\mbf A}\sigma(\bfet),
\end{equation} whence 
\begin{equation}
\dot{{\mbf r}} = -\frac{1}{\tau}{\mbf r} + {\mbf B}{\mbf A}\sigma(\bfet) + \frac{1}{\tau} \bm{\theta}.
\end{equation} Thus, if ${\mbf r} \approx \bfet$ then 
\begin{equation}
\dot{{\mbf r}} - \dot{\bfet} \approx \frac{1}{\tau} \bm{\theta}.
\label{app}
\end{equation}
Recall that ${\mbf r} \approx \bfet$ is exact at $t=0$. Integrating~\ref{app}, then taking magnitudes and deriving yields
\begin{equation}
\dot{\delta} = \frac{d}{dt}||{\mbf r} - \bfet|| = \frac{1}{\tau}||\bm{\theta}||.
\label{res}
\end{equation}
Therefore, when $\tau$ is large $\dot{\delta}$ is small. This is the desired result.

To verify~\ref{res}, we have computed $\delta(t)$ for various trajectories in synthesized RNNs. For example, in the two-attractor system depicted in Figure~\ref{fig:2fps}, it happens that $\tau^{-1}||\bm{\theta}|| = 4.7756\times10^{-6}$. W measured the evolution of $\delta(t)$ along one of the depicted orbits and found that it is a linear function of time whose slope is $4.7757\times10^{-6}$, agreeing almost perfectly with $\tau^{-1}||\bm{\theta}||$. This has been the case for all orbits we have measured.



\subsection{Potential energy functions for recurrent networks}
\cite{mendes1992} showed that the additive recurrent neural networks can be decomposed as the sum of a gradient system and a Hamiltonian system as follows:
\begin{equation}
\dot{{\mbf s}} = {\mbf P}({\mbf s})\nabla V({\mbf s}) + {\mbf Q}({\mbf s})\nabla H({\mbf s}) + {\mbf E}{\mbf u}(t).
\end{equation}
The scalar functions $V$ and $H$ are the gradient system's potential function and the Hamiltonian system's potential function, respectively. The third term represents inputs to the network.

The appropriate gradient potential for this decomposition is given by
\begin{equation}
V({\mbf s}) = \sum_{i=1}^n \int^{s_i}a_i(\zeta_i) \zeta_i f_i'(\zeta_i) d\zeta_i - \frac{1}{2}\sum_{i=1}^n \sum_{j=1}^n w_{ij}^S f_i(s_i) f_j(s_j)
\label{gradient}
\end{equation}
where $a_i(s_i)$ is the refractory parameter (in our networks $a_i = \tau^{-1}$), $f_i(s_i)$ is the neuron activation function (in our networks $f_i(s_i) = \sigma(s_i)~\forall i$), $s_i$ is the state of the $i$th neuron, and $w_{ij}^S$ is the $(i,j)$-entry in the symmetric decomposition of the network's weight matrix.

Here we can recognize a symptom of the space mismatch inherent to dynamical approximation by recurrent neural networks. The system being approximated is $n$-dimensional and its potential function lies in $n$-dimensional space. On the other hand, the gradient potential~\ref{gradient} depends on $n+m$ neurons. We can determine potentials along integrated trajectories, but to determine potentials over the entire output space in this manner requires a great deal of computation.

\begin{figure}
  \centering
  \includegraphics[width=4in]{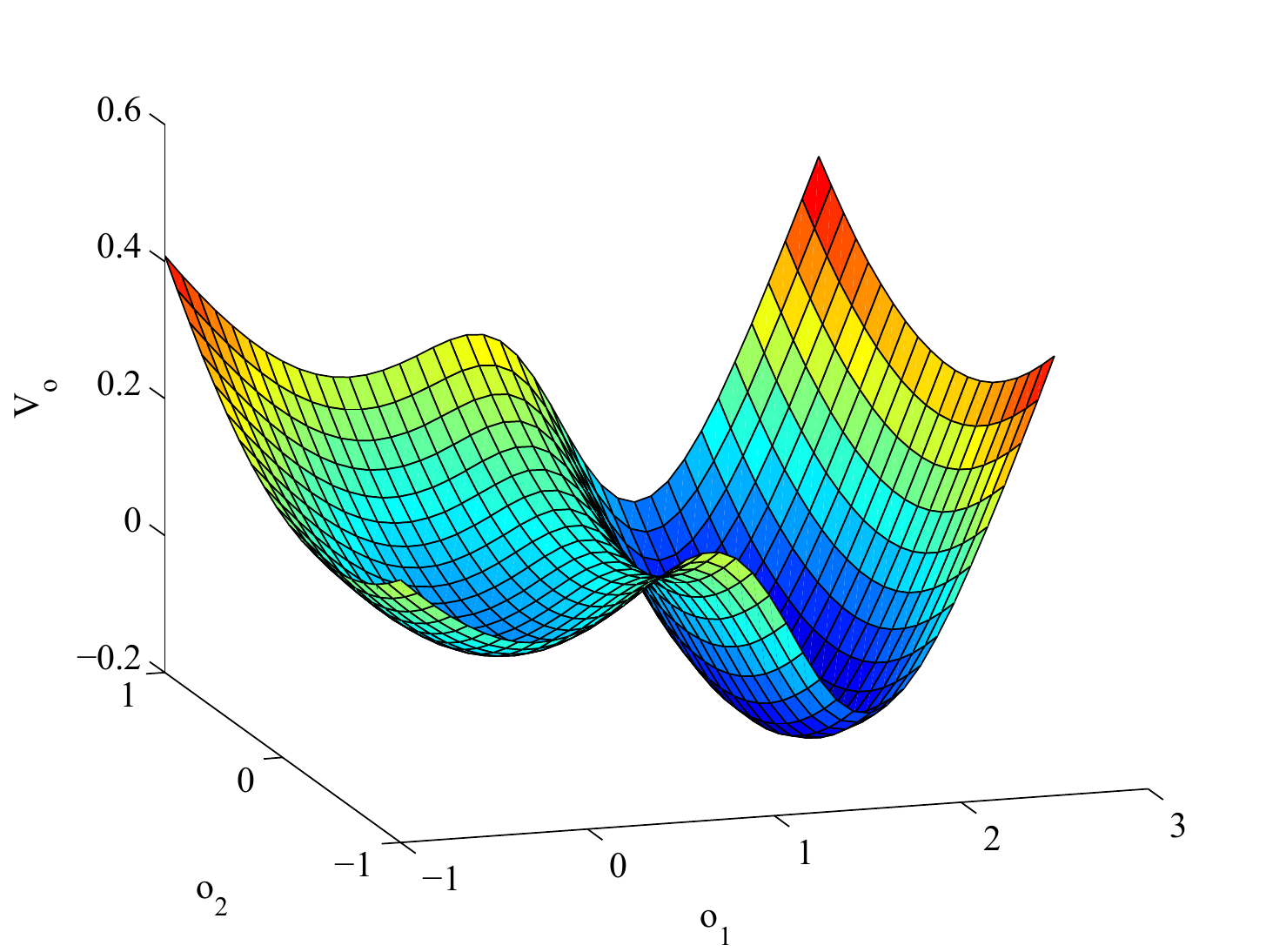}
  \caption{ Gradient potential for the system depicted in Figure~\ref{fig:2fps}(a).}
  \label{fig:pot1}
\end{figure}

It is also possible to use the initial condition, whose hidden component we now recognize as the approximator ${\mbf r}(t)$. By the result of the previous subsection, this expression holds (approximately) true throughout the course of a trajectory if $\tau$ is large. Then, since ${\mbf r}$ is a function of $\bfom$, to close approximation the gradient potential can be represented as a function of the output states. When $\tau$ is large we can also simplify~\ref{gradient} by neglecting the first term. The approximate gradient potential is then given by
\begin{equation}
V_{\bfom} ({\mbf y}) \triangleq - \frac{1}{2}\sum_{i=1}^n \sum_{j=1}^n w_{ij}^s \sigma_i(y_i) \sigma_j(y_j),
\label{appgrad}
\end{equation}
where ${\mbf y} = \text{col}[\bfom, {\mbf r}] = \text{col}[\bfom, {\mbf B}\bfom+\bm{\theta}]$. In Figure~\ref{fig:pot1} we plot this approximate gradient for the two-attractor system of Figure~\ref{fig:2fps}. The two basins of attraction and the saddle that separates them are clearly visible.

What is perhaps most interesting about the foregoing is that we can visualize the dynamics of a network (using the gradient potential) without solving a single differential equation, and based solely on the network weights. This could have interesting applications, for instance in estimating the capacity of attractor-based memory networks.

\subsection{Relation to the Neural Engineering Framework}
\label{sub:nef}
We now make explicit the analogies between our RNN training procedure and the Neural Engineering Framework (NEF).

Both the NEF and our framework use a vector-field perspective in their modeling of neuron activity \citep{eliasmith2005}. In the NEF, neuron activities $a_i$ encoding some vector ${\mbf q}$ are given by
\begin{equation*}
a_i({\mbf q}) = G_i[\alpha_i \langle{\mbf q} \cdot \tilde{\phi_i}\rangle + J_i^\text{bias}],
\end{equation*}
where $G_i$ is the nonlinear activation function, $\alpha_i$ is a gain and conversion factor, $J_i^\text{bias}$ is a current signal that accounts for background activity, and $\tilde{\phi_i}$ is the linear encoding weight. A transformation $F({\mbf q})$ on the variable ${\mbf q}$ can be performed according to
\begin{equation*}
\hat{F}({\mbf q}) = \sum_i a_i({\mbf q}) \phi_i^{F({\mbf q})},
\end{equation*}
where $\phi_i^{F({\mbf q})}$ is a linear decoding weight and $\hat{F}$ is an approximation of the desired transformation.

The linear encoders $\tilde{\phi_i}$ of the NEF correspond to the first-layer weights ${\mbf B}$ in our feedforward networks, while the linear decoders $\phi_i^{F({\mbf q})}$ correspond to the second-layer weights ${\mbf A}$. We can make this clear by rewriting the NEF equations in matrix-vector form. For the neuron activities we have
\begin{equation*}
{\mbf a} = \bm{G}(\tilde{\bm{\Phi}}{\mbf q}),
\end{equation*}
where $\tilde{\bm{\Phi}}$ is the matrix whose rows are given by the scaled encoding vectors $\alpha_i \tilde{\phi}_i$, $\bm{G}$ is the vector of activation functions, and we ignore the input current $J^\text{bias}$. For the decoding stage we have 
\begin{equation*}
\hat{F}({\mbf q}) = \bm{\Phi}^{F({\mbf q})} {\mbf a},
\end{equation*}
where now $\bm{\Phi}^{F({\mbf q})}$ is the matrix whose columns are given by the decoding vectors $\phi_i^{F({\mbf q})}$. This makes it obvious that ${\mbf B}$ plays the role of $\tilde{\bm{\Phi}}$ while ${\mbf A}$ plays the role of $\bm{\Phi}^{F({\mbf q})}$.

Our method diverges from the NEF in the training procedure. In the NEF, the linear encoders are set randomly, while the set of linear decoders is computed by minimizing the mean squared error at the output layer only. On the other hand, our training methodology uses backpropagation to train {\it both} sets of weights. Because errors are propagated through the entire model, it is possible that systems may be modeled more accurately using fewer neurons with our approach, although a direct comparison has not been undertaken. To our knowledge, it was not known before this work that the theoretical bounds of~\cite{funahashi1993} could be applied to this restricted class of NEF systems.

\subsection{Relation to reservoir computing}
\label{sub:d2}
There also exists some structural overlap between the RNNs constructed by our algorithm and reservoir computers.

The key feature of reservoir computers is the set of recurrently connected neurons whose weights are set randomly and then frozen. The reservoir connects to a set of output units. Reservoir units, whose states we collect in vector ${\mbf x}(t) \in {}^m\mathbb{R}$, typically operate by sigmoidal activation functions, while output units, whose states we collect in vector ${\mbf y}(t) \in {}^n\mathbb{R}$, operate by linear activation functions. Output states are determined from reservoir states as ${\mbf y}(t) = {\mbf Q}{\mbf x}(t)$. Note that this is an algebraic rather than a differential relation.

A reservoir computer's output units may be made to approximate prescribed dynamical orbits through a unique training approach. Typically, only the output weights ${\mbf Q}$ that feed the output units are trained ({\it e.g.}, by linear regression or recursive least squares~\citep{sussillo2009}). The random recurrent connections ${\mbf R}$ within the reservoir remain fixed during this training.\footnote{Note the similarity to the NEF training procedure.} The reservoir may or may not have external inputs, but it features feedback connections from the output units to the reservoir; these connections are also randomly initialized and fixed throughout training.

It should be clear that the hidden neurons $\bfet(t)$ in our networks, which undergo sigmoidal activation and are connected recurrently by matrix ${\mbf B}{\mbf A}$, act like a reservoir; that is, $\bfet(t)$ is analogous to  ${\mbf x}(t)$ and ${\mbf B}{\mbf A}$ is analogous to ${\mbf R}$.

This analogy suggests that recurrently connected hidden-unit sets, trained by our algorithm, can be used as continuous-time reservoirs. Our feedforward training can be used to generate reservoirs that are tailored to desired dynamical regimes, such as limit cycles or strange attractors. Our training algorithm and the attendant vector field perspective may thus shed light on some of the unanswered questions regarding reservoir properties and network dynamics (see \cite{jaeger2009}).

\section{Concluding Remarks}
\label{sec:conc}
In this work we introduced a novel algorithm for the synthesis of recurrent neural networks that implement general dynamical systems. In this algorithm, a three-layer feedforward neural network given by ${\mbf A}\bm{\sigma}({\mbf B}{\mbf q}+\bm{\theta})$ is trained to approximate an arbitrary dynamical system given by the vector field $\dot{{\mbf q}} = F({\mbf q})$, and then transformed into a recurrent neural network based on a theorem from \cite{funahashi1993}. The recurrent network so constructed closely reproduces the original system's dynamics as reported in Section~\ref{sec:res}.

Our algorithm represents a new tool for the synthesis of continuous-time recurrent neural networks. These are a large, broadly applicable class of models with a structural analogy to the brain. It is hoped that this tool can be used to design and implement brain-like systems for artificial intelligence and neuroscience research, and, particularly, to make testable predictions about neural organization in living systems.

\section{Acknowledgment}
We thank the reviewers and the editor for their illuminating and insightful comments, which have helped to improve this paper.

\section*{References}

\bibliography{NNrefs}

\begin{thebibliography}{29}
\expandafter\ifx\csname natexlab\endcsname\relax\def\natexlab#1{#1}\fi
\providecommand{\url}[1]{\texttt{#1}}
\providecommand{\href}[2]{#2}
\providecommand{\path}[1]{#1}
\providecommand{\DOIprefix}{doi:}
\providecommand{\ArXivprefix}{arXiv:}
\providecommand{\URLprefix}{URL: }
\providecommand{\Pubmedprefix}{pmid:}
\providecommand{\doi}[1]{\href{http://dx.doi.org/#1}{\path{#1}}}
\providecommand{\Pubmed}[1]{\href{pmid:#1}{\path{#1}}}
\providecommand{\bibinfo}[2]{#2}
\ifx\xfnm\relax \def\xfnm[#1]{\unskip,\space#1}\fi
\bibitem[{Afraimovich et~al.(2004)Afraimovich, Rabinovich \&
  Varona}]{afraimovich2004}
\bibinfo{author}{Afraimovich, V.~S.}, \bibinfo{author}{Rabinovich, M.~I.}, \&
  \bibinfo{author}{Varona, P.} (\bibinfo{year}{2004}).
\newblock \bibinfo{title}{Heteroclinic contours in neural ensembles and the
  winnerless competition principle}.
\newblock {\it \bibinfo{journal}{International Journal of Bifurcation and
  Chaos}\/},  {\it \bibinfo{volume}{14}\/}, \bibinfo{pages}{1195--1208}.
\bibitem[{Amit(1992)}]{amit1992}
\bibinfo{author}{Amit, D.~J.} (\bibinfo{year}{1992}).
\newblock {\it \bibinfo{title}{Modeling brain function: The world of attractor
  neural networks}\/}.
\newblock \bibinfo{publisher}{Cambridge University Press}.
\bibitem[{Bao \& Zeng(2012)}]{bao2012}
\bibinfo{author}{Bao, G.}, \& \bibinfo{author}{Zeng, Z.}
  (\bibinfo{year}{2012}).
\newblock \bibinfo{title}{Analysis and design of associative memories based on
  recurrent neural network with discontinuous activation functions}.
\newblock {\it \bibinfo{journal}{Neurocomputing}\/},  {\it
  \bibinfo{volume}{77}\/}, \bibinfo{pages}{101--107}.
\bibitem[{Bottou(2010)}]{bottou2010}
\bibinfo{author}{Bottou, L.} (\bibinfo{year}{2010}).
\newblock \bibinfo{title}{Large-scale machine learning with stochastic gradient
  descent}.
\newblock In {\it \bibinfo{booktitle}{Proceedings of COMPSTAT'2010}\/} (pp.
  \bibinfo{pages}{177--186}).
\newblock \bibinfo{publisher}{Springer}.
\bibitem[{Chow \& Li(2000)}]{chow2000}
\bibinfo{author}{Chow, T.~W.}, \& \bibinfo{author}{Li, X.-D.}
  (\bibinfo{year}{2000}).
\newblock \bibinfo{title}{Modeling of continuous time dynamical systems with
  input by recurrent neural networks}.
\newblock {\it \bibinfo{journal}{Circuits and Systems I: Fundamental Theory and
  Applications, IEEE Transactions on}\/},  {\it \bibinfo{volume}{47}\/},
  \bibinfo{pages}{575--578}.
\bibitem[{Eliasmith(2005)}]{eliasmith2005}
\bibinfo{author}{Eliasmith, C.} (\bibinfo{year}{2005}).
\newblock \bibinfo{title}{A unified approach to building and controlling
  spiking attractor networks}.
\newblock {\it \bibinfo{journal}{Neural computation}\/},  {\it
  \bibinfo{volume}{17}\/}, \bibinfo{pages}{1276--1314}.
\bibitem[{Eliasmith \& Anderson(2004)}]{eliasmith2004}
\bibinfo{author}{Eliasmith, C.}, \& \bibinfo{author}{Anderson, C. C.~H.}
  (\bibinfo{year}{2004}).
\newblock {\it \bibinfo{title}{Neural engineering: Computation, representation,
  and dynamics in neurobiological systems}\/}.
\newblock \bibinfo{publisher}{MIT Press}.
\bibitem[{Feldkamp et~al.(1998)Feldkamp, Prokhorov, Eagen \&
  Yuan}]{feldkamp1998}
\bibinfo{author}{Feldkamp, L.~A.}, \bibinfo{author}{Prokhorov, D.~V.},
  \bibinfo{author}{Eagen, C.~F.}, \& \bibinfo{author}{Yuan, F.}
  (\bibinfo{year}{1998}).
\newblock \bibinfo{title}{Enhanced multi-stream kalman filter training for
  recurrent networks}.
\newblock In {\it \bibinfo{booktitle}{Nonlinear Modeling}\/} (pp.
  \bibinfo{pages}{29--53}).
\newblock \bibinfo{publisher}{Springer}.
\bibitem[{Funahashi(1989)}]{funahashi1989}
\bibinfo{author}{Funahashi, K.-I.} (\bibinfo{year}{1989}).
\newblock \bibinfo{title}{On the approximate realization of continuous mappings
  by neural networks}.
\newblock {\it \bibinfo{journal}{Neural networks}\/},  {\it
  \bibinfo{volume}{2}\/}, \bibinfo{pages}{183--192}.
\bibitem[{Funahashi \& Nakamura(1993)}]{funahashi1993}
\bibinfo{author}{Funahashi, K.-I.}, \& \bibinfo{author}{Nakamura, Y.}
  (\bibinfo{year}{1993}).
\newblock \bibinfo{title}{Approximation of dynamical systems by continuous time
  recurrent neural networks}.
\newblock {\it \bibinfo{journal}{Neural networks}\/},  {\it
  \bibinfo{volume}{6}\/}, \bibinfo{pages}{801--806}.
\bibitem[{Hermans \& Schrauwen(2010)}]{hermans2010}
\bibinfo{author}{Hermans, M.}, \& \bibinfo{author}{Schrauwen, B.}
  (\bibinfo{year}{2010}).
\newblock \bibinfo{title}{Memory in linear recurrent neural networks in
  continuous time}.
\newblock {\it \bibinfo{journal}{Neural Networks}\/},  {\it
  \bibinfo{volume}{23}\/}, \bibinfo{pages}{341--355}.
\bibitem[{Hochreiter \& Schmidhuber(1997)}]{hochreiter1997}
\bibinfo{author}{Hochreiter, S.}, \& \bibinfo{author}{Schmidhuber, J.}
  (\bibinfo{year}{1997}).
\newblock \bibinfo{title}{Long short-term memory}.
\newblock {\it \bibinfo{journal}{Neural Computation}\/},  {\it
  \bibinfo{volume}{9}\/}, \bibinfo{pages}{1735--1780}.
\bibitem[{Hooper(2001)}]{hooper2001}
\bibinfo{author}{Hooper, S.~L.} (\bibinfo{year}{2001}).
\newblock \bibinfo{title}{Central pattern generators}.
\newblock {\it \bibinfo{journal}{eLS}\/}, .
\bibitem[{Hopfield(1982)}]{hopfield1982}
\bibinfo{author}{Hopfield, J.~J.} (\bibinfo{year}{1982}).
\newblock \bibinfo{title}{Neural networks and physical systems with emergent
  collective computational abilities}.
\newblock {\it \bibinfo{journal}{Proceedings of the national academy of
  sciences}\/},  {\it \bibinfo{volume}{79}\/}, \bibinfo{pages}{2554--2558}.
\bibitem[{Jaeger \& Haas(2004)}]{jaeger2004}
\bibinfo{author}{Jaeger, H.}, \& \bibinfo{author}{Haas, H.}
  (\bibinfo{year}{2004}).
\newblock \bibinfo{title}{Harnessing nonlinearity: Predicting chaotic systems
  and saving energy in wireless communication}.
\newblock {\it \bibinfo{journal}{Science}\/},  {\it \bibinfo{volume}{304}\/},
  \bibinfo{pages}{78--80}.
\bibitem[{Kaneko \& Tsuda(2003)}]{kaneko2003}
\bibinfo{author}{Kaneko, K.}, \& \bibinfo{author}{Tsuda, I.}
  (\bibinfo{year}{2003}).
\newblock \bibinfo{title}{Chaotic itinerancy}.
\newblock {\it \bibinfo{journal}{Chaos: An Interdisciplinary Journal of
  Nonlinear Science}\/},  {\it \bibinfo{volume}{13}\/},
  \bibinfo{pages}{926--936}.
\bibitem[{Kingma \& Ba(2014)}]{kingma2014}
\bibinfo{author}{Kingma, D.}, \& \bibinfo{author}{Ba, J.}
  (\bibinfo{year}{2014}).
\newblock \bibinfo{title}{Adam: A method for stochastic optimization}.
\newblock {\it \bibinfo{journal}{arXiv preprint arXiv:1412.6980}\/}, .
\bibitem[{Luko{\v{s}}evi{\v{c}}ius \& Jaeger(2009)}]{jaeger2009}
\bibinfo{author}{Luko{\v{s}}evi{\v{c}}ius, M.}, \& \bibinfo{author}{Jaeger, H.}
  (\bibinfo{year}{2009}).
\newblock \bibinfo{title}{Survey: Reservoir computing approaches to recurrent
  neural network training}.
\newblock {\it \bibinfo{journal}{Computer Science Review}\/},  {\it
  \bibinfo{volume}{3}\/}, \bibinfo{pages}{127--149}.
\bibitem[{Martens \& Sutskever(2011)}]{martens2011}
\bibinfo{author}{Martens, J.}, \& \bibinfo{author}{Sutskever, I.}
  (\bibinfo{year}{2011}).
\newblock \bibinfo{title}{Learning recurrent neural networks with hessian-free
  optimization}.
\newblock In {\it \bibinfo{booktitle}{Proceedings of the 28th International
  Conference on Machine Learning (ICML-11)}\/} (pp.
  \bibinfo{pages}{1033--1040}).
\bibitem[{Mendes \& Duarte(1992)}]{mendes1992}
\bibinfo{author}{Mendes, R.~V.}, \& \bibinfo{author}{Duarte, J.~T.}
  (\bibinfo{year}{1992}).
\newblock \bibinfo{title}{Vector fields and neural networks}.
\newblock {\it \bibinfo{journal}{Complex Systems}\/},  {\it
  \bibinfo{volume}{6}\/}, \bibinfo{pages}{21--30}.
\bibitem[{Rumelhart et~al.(1988)Rumelhart, Hinton \& Williams}]{rumelhart1988}
\bibinfo{author}{Rumelhart, D.~E.}, \bibinfo{author}{Hinton, G.~E.}, \&
  \bibinfo{author}{Williams, R.~J.} (\bibinfo{year}{1988}).
\newblock \bibinfo{title}{Learning representations by back-propagating errors}.
\newblock {\it \bibinfo{journal}{Cognitive modeling}\/},  {\it
  \bibinfo{volume}{5}\/}, \bibinfo{pages}{3}.
\bibitem[{Sotomayor \& Teixeira(1996)}]{sotomayor1996}
\bibinfo{author}{Sotomayor, J.}, \& \bibinfo{author}{Teixeira, M.}
  (\bibinfo{year}{1996}).
\newblock \bibinfo{title}{Regularization of discontinuous vector fields}.
\newblock In {\it \bibinfo{booktitle}{International Conference on Differential
  Equations, Lisboa}\/} (pp. \bibinfo{pages}{207--223}).
\bibitem[{Sussillo \& Abbott(2009)}]{sussillo2009}
\bibinfo{author}{Sussillo, D.}, \& \bibinfo{author}{Abbott, L.~F.}
  (\bibinfo{year}{2009}).
\newblock \bibinfo{title}{Generating coherent patterns of activity from chaotic
  neural networks}.
\newblock {\it \bibinfo{journal}{Neuron}\/},  {\it \bibinfo{volume}{63}\/},
  \bibinfo{pages}{544--557}.
\bibitem[{Sutskever et~al.(2013)Sutskever, Martens, Dahl \&
  Hinton}]{sutskever2013}
\bibinfo{author}{Sutskever, I.}, \bibinfo{author}{Martens, J.},
  \bibinfo{author}{Dahl, G.}, \& \bibinfo{author}{Hinton, G.}
  (\bibinfo{year}{2013}).
\newblock \bibinfo{title}{On the importance of initialization and momentum in
  deep learning}.
\newblock In {\it \bibinfo{booktitle}{Proceedings of the 30th international
  conference on machine learning (ICML-13)}\/} (pp.
  \bibinfo{pages}{1139--1147}).
\bibitem[{Takens(1981)}]{takens1981}
\bibinfo{author}{Takens, F.} (\bibinfo{year}{1981}).
\newblock {\it \bibinfo{title}{Detecting strange attractors in turbulence}\/}.
\newblock \bibinfo{publisher}{Springer}.
\bibitem[{Trischler \& D'Eleuterio(2013)}]{trischler2013}
\bibinfo{author}{Trischler, A.~P.}, \& \bibinfo{author}{D'Eleuterio, G.~M.}
  (\bibinfo{year}{2013}).
\newblock \bibinfo{title}{Sculpting dynamical systems for models of neural
  computation and memory}.
\newblock {\it \bibinfo{journal}{BMC Neuroscience}\/},  {\it
  \bibinfo{volume}{14}\/}, \bibinfo{pages}{P103}.
\bibitem[{Tsung \& Cottrell(1995)}]{tsung1995}
\bibinfo{author}{Tsung, F.-S.}, \& \bibinfo{author}{Cottrell, G.~W.}
  (\bibinfo{year}{1995}).
\newblock \bibinfo{title}{Phase-space learning}.
\newblock {\it \bibinfo{journal}{Advances in Neural Information Processing
  Systems}\/},  (pp. \bibinfo{pages}{481--488}).
\bibitem[{Werbos(1990)}]{werbos1990}
\bibinfo{author}{Werbos, P.~J.} (\bibinfo{year}{1990}).
\newblock \bibinfo{title}{Backpropagation through time: what it does and how to
  do it}.
\newblock {\it \bibinfo{journal}{Proceedings of the IEEE}\/},  {\it
  \bibinfo{volume}{78}\/}, \bibinfo{pages}{1550--1560}.
\bibitem[{Williams \& Zipser(1989)}]{williams1989}
\bibinfo{author}{Williams, R.~J.}, \& \bibinfo{author}{Zipser, D.}
  (\bibinfo{year}{1989}).
\newblock \bibinfo{title}{A learning algorithm for continually running fully
  recurrent neural networks}.
\newblock {\it \bibinfo{journal}{Neural computation}\/},  {\it
  \bibinfo{volume}{1}\/}, \bibinfo{pages}{270--280}.

\end{thebibliography}

\section{Appendix}
We prove Theorem~\ref{thm3} for the forced recurrent neural network, following the path of \cite{funahashi1993}.

\begin{pf}
Recall that the system $\xdot = R(\bfx) + \bfE\bfx$ accounts for the dynamics of the forcing function, which enters additively via the $\bfE\bfx$ term. We define $\Rtilde(\bfx)$ as
\begin{equation}
\Rtilde(\bfx) \triangleq -\frac{1}{\tau}\bfx + \bfA\bm{\sigma}(\bfB\bfx + \bm{\theta}).
\end{equation}
Provided that we can determine the required $\bfA$, $\bfB$, and $\bm{\theta}$ such that
\begin{equation}
    \max_\bfx \| R(\bfx) - {\mbf A}\bm{\sigma}({\mbf B}\bfx+\bm{\theta}) \| < \frac{\epsilon L_R}{4(e^{L_R T}-1)},
\label{eq:one}
\end{equation}
holds, and conditions similar to \ref{eq:tau} are met, then
\begin{equation}
    \max_\bfx \|R(\bfx) - \Rtilde(\bfx)\| \leq \frac{\epsilon L_R}{4(e^{L_R T} - 1)}
\end{equation}
by~\cite{funahashi1993}. It follows that
\begin{equation}
    \max_\bfx \|R(\bfx) + \bfE\bfx - (\Rtilde(\bfx) + \bfE\bfx)\| 
        \leq \frac{\epsilon L_R}{4(e^{L_R T} - 1)},
\end{equation}
which in turn implies
\begin{equation}
    \max_{t\in I} \|\bfx(t) - \xtilde(t)\| \leq \frac{\epsilon}{2}
\label{bound:1}
\end{equation}
where $\xtilde$ is the solution to $\dot{\xtilde} = \Rtilde(\xtilde) + \bfE\xtilde$. Note that $\epsilon$ can be arbitrarily specified.
 
As in the unforced case, we build $S$ on the blueprint of $G$ and $\Stilde$ as
\begin{equation}
	\Stilde(\bfz) \triangleq -\frac{1}{\tau}\bfz + {\mbf W}\bm{\sigma}(\bfz) + \frac{1}{\tau}\bm{\theta},
\end{equation}
identifying the overall state as $\bfz$ in the place of ${\mbf s}$.  Now
\begin{equation}
    \|S(\bfz) - \Stilde(\bfz)\| < \frac{\epsilon L_\Stilde}{4(e^{L_\Stilde T} - 1)}
\end{equation}
whence we also have that
\begin{equation}
    \|S(\bfz) + \bfH\bfz - (\Stilde(\bfz) + \bfH\bfz)\| 
        < \frac{\epsilon L_\Stilde}{4(e^{L_\Stilde T} - 1)}
\end{equation}
where $\bfH \triangleq \text{col}\,[\bfE, \Zero]$. Thus
\begin{equation}
    \max_{t\in I} \|\ztilde(t) - \bfz(t)\| \leq \frac{\epsilon}{2}
\label{bound:2}
\end{equation}
where $\bfz$ is the solution to $\dot{\bfz} = S(\bfz) + \bfH\bfz$ and $\ztilde$ the solution to
\begin{equation}
    \dot{\ztilde} = \Stilde(\ztilde) + \bfH\ztilde
\end{equation}
which can be realized as a forced recurrent neural network.  Inequality \ref{bound:2} holds for any subset of $\bfz$ and its companion subset of $\ztilde$; in particular,
\begin{equation}
    \max_{t\in I} \|\xtilde(t) - {\bm\varpi}(t)\| \leq \frac{\epsilon}{2}
\label{bound:3}
\end{equation}
where $\bm{\varpi}(t)$ is the state of output neurons of the FRNN corresponding to $\xtilde$, which we recall is the solution to $\dot{\xtilde} = \Rtilde(\xtilde) + \bfE\xtilde$.

Combining \ref{bound:1} and \ref{bound:3}, we conclude that 
\begin{equation}    
    \max_{t\in I} \|\bfx(t) - {\bm\varpi}(t)\| \leq \epsilon
\end{equation}
For the problem at hand, we recognize that $\bfq(t)$, the solution to the true forced dynamical system, would be a subset of states in $\bfx(t)$ and thus
\begin{equation}
    \max_{t\in I} \|\bfq(t) - {\bm\omega}(t)\| \leq \epsilon
\end{equation}
where $\bfom(t)$ represents the states of the corresponding subset of the output FRNN neurons.

\end{pf}

\end{document}